\documentclass{article}

\PassOptionsToPackage{square, numbers}{natbib}

\usepackage[preprint]{neurips_2024}

\usepackage[utf8]{inputenc} 
\usepackage[T1]{fontenc}    
\usepackage{url}            
\usepackage{booktabs}       
\usepackage{amsfonts}       
\usepackage{nicefrac}       
\usepackage{microtype}      
\usepackage{xcolor}         
\usepackage{tabularx}
\usepackage{subcaption}
\usepackage{multicol}
\usepackage{multirow}
\usepackage{tabularray}
\usepackage{amsmath}
\usepackage{amssymb}
\usepackage{siunitx}
\usepackage{colortbl}
\usepackage{graphicx}
\usepackage{lipsum}
\usepackage{hyperref}       
\usepackage[capitalize]{cleveref}

\usepackage{pgfplots,gincltex}
\usepgfplotslibrary{groupplots}
\pgfplotsset{compat=1.6}
\usepackage{tikz}
\usetikzlibrary{arrows.meta}

\definecolor{cvprblue}{rgb}{0.21,0.49,0.74}
\definecolor{pltblue}{RGB}{174, 199, 232}
\definecolor{pltorange}{RGB}{255, 229, 204}
\definecolor{pltgreen}{RGB}{204, 229, 204}
\definecolor{pltred}{RGB}{229, 204, 204}
\definecolor{pltpurple}{RGB}{239, 218, 230}

\definecolor{tabblue}{HTML}{1f77b4}
\definecolor{taborange}{HTML}{ff7f0e}
\definecolor{tabgreen}{HTML}{2ca02c}
\definecolor{tabred}{HTML}{d62728}
\definecolor{tabpurple}{HTML}{9467bd}

\definecolor{cblue}{RGB}{173, 201, 233}
\definecolor{clblue}{RGB}{222, 234, 246}
\definecolor{corange}{RGB}{255, 152, 67}
\definecolor{lorgange}{RGB}{255, 221, 149}

\newcolumntype{C}{>{\centering\arraybackslash}X}

\title{Efficient Degradation-aware Any Image Restoration}

\author{%
  Eduard Zamfir$^{1}$ \quad Zongwei Wu$^{1}$ \quad Nancy Mehta$^{1}$ \\
  \textbf{Danda Pani Paudel}$^{2,3}$ \quad \textbf{Yulun Zhang}$^{4}$ \quad \textbf{Radu Timofte}$^{1}$ \\
  $^{1}$University of Würzburg \quad $^{2}$INSAIT Sofia University \\
  $^{3}$ ETH Zürich \quad $^{4}$Shanghai Jiao Tong University\\
}

\begin{document}

\maketitle

\begin{abstract}
Reconstructing missing details from degraded low-quality inputs poses a significant challenge. Recent progress in image restoration has demonstrated the efficacy of learning large models capable of addressing various degradations simultaneously. 
Nonetheless, these approaches introduce considerable computational overhead and complex learning paradigms, limiting their practical utility. 
In response, we propose \textit{DaAIR}, an efficient All-in-One image restorer employing a Degradation-aware Learner (DaLe) in the low-rank regime to collaboratively mine shared aspects and subtle nuances across diverse degradations, generating a degradation-aware embedding. By dynamically allocating model capacity to input degradations, we realize an efficient restorer integrating holistic and specific learning within a unified model. 
Furthermore, DaAIR introduces a cost-efficient parameter update mechanism that enhances degradation awareness while maintaining computational efficiency.
Extensive comparisons across five image degradations demonstrate that our DaAIR outperforms both state-of-the-art All-in-One models and degradation-specific counterparts, affirming our efficacy and practicality. The source will be publicly made available at \url{https://eduardzamfir.github.io/daair/}

\end{abstract}
\section{Introduction}

Image restoration is a fundamental problem in computer vision, focusing on reconstructing high-quality images from deteriorated observations. Adverse conditions such as noise, haze, or rain significantly impact the practical utility of images in downstream tasks across various domains, including autonomous navigation~\cite{valanarasu2022transweather,chen2023alwayscleardays} and augmented reality\cite{girbacia2013virtual,saggio2011augmented,dang2020application}. Therefore, developing robust image restoration techniques is a critical endeavour.
Recent advances in Deep learning-based approaches have shown great achievements in image restoration~\cite{zhang2017learning,tai2017memnet,lehtinen2018noise2noise,zhang2019residual,liu2021swin,chen2022simple, Zamir2021Restormer,chen2022cross,li2023pip,cui2024adair,dudhane2024dynet}. However, most of the existing works adopt task-specific learning, targeting a single known degradation at once ~\cite{zhang2018residual,liu2021swin,Zamir2021Restormer,chen2022cross,chen2022simple}. 
This specificity, of one model per task, inherently limits their practicability and hinders their application in diverse degradation settings~\cite{zamir2020cycleisp,zamir2020learning,purohit2021spatially}. 

To address this limitation, there has been growing interest in All-in-One image restoration models capable of handling various degradations simultaneously. Notable works in this direction ~\cite{li2022airnet,zamir2021pmrnet,valanarasu2022transweather,liu2022tape,fan2019dl,potlapalli2023promptir,zhang2023ingredient} employ contrastive~\cite{li2022airnet}, meta-learning~\cite{zhang2023ingredient} or visual prompting techniques~\cite{potlapalli2023promptir,wang2023promptrestorer}. Despite their success, these models face two primary limitations. 
First, these models often overlook the distinct and shared characteristics of each image corruption. Even when certain approaches ~\cite{liu2022tape,zhang2023ingredient} account for these characteristics, they often rely on external prior information or complex progressive meta-learning, thus failing to harness the inherent potential benefits of self-learning within the network. Second, leveraging auxiliary priors through prompt-learning~\cite{potlapalli2023promptir,wang2023promptrestorer,li2023pip,dudhane2024dynet} is a prominent approach. However, their task-specific relevance may be ambiguous, necessitating abundant learnable parameters to capture intricate details, complicating the model interpretability and increasing the overall computational demands. 

Unlike other works, our objective is to recognize degradation-specific representations while emphasizing the commonalities that unify distinct degradations into a shared underlying representation, leveraging the intricacies among different restoration tasks efficiently. 
For instance, dehazing and denoising have traditionally been considered separate tasks. Haze manifests as a uniform veil that reduces contrast and makes objects appear less distinct, while noise appears as random variations in pixel values, often resembling graininess. Despite their distinct visual effects, these degradations share similar characteristics: both haze and noise are distributed across the entire image and affect similar aspects, such as edges, textures, and contrasts. Thus, learning how to dehaze should also contribute to denoising tasks. Further, we foresee that such associations are also prevalent in various other degradations. 

\begin{figure}[t]
    \begin{minipage}[b]{0.49\textwidth}
        \centering
        \begin{subfigure}{0.48\textwidth}
             \includegraphics[width=\textwidth]{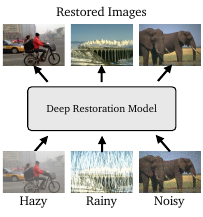}
             \subcaption{}
        \end{subfigure}
        \hfill
        \begin{subfigure}{0.48\textwidth}
            \includegraphics[width=\textwidth]{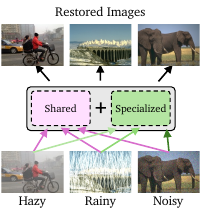}
            \subcaption{}
        \end{subfigure}
        \caption{(a) Prior work employ inefficient models and learning for capturing degradation dependencies. (b) We route model capacity to efficiently learn shared and specialized embeddings.}
        \label{fig:motivation}
    \end{minipage}
    \hfill
    \begin{minipage}[b]{0.49\textwidth}
        \centering
        \begin{tikzpicture}[baseline]

\begin{axis}[
    width=\textwidth,
    height=0.65\textwidth,
    xtick pos=left,
    ytick pos=left,
    xlabel near ticks,
    ylabel near ticks, 
    ymin = 25,
    ymax = 33,
    xmin = 0,
    xmax = 50,
    xlabel={Parameters (M)},
    ylabel={PSNR (dB)},
    grid,
    grid style=dashed,
    title style={anchor=north, at={(0.5,1.1)},},
    tick label style={font=\scriptsize}, 
    title style={font=\scriptsize}, 
    label style={font=\scriptsize}, 
    nodes near coords, 
    enlarge y limits=0.1,
    enlarge x limits=0.1,
    legend style={at={(.2,.925)}, anchor=center, legend columns=-1, font=\tiny,},
    ]
    
    \addplot[
    nodes near coords style = {font=\scriptsize, anchor=south west, xshift=-2pt},
    scatter/classes={
            b={color=tabblue!50, mark size=1.7},
            c={color=taborange!50, mark size=1.7}
            },
    scatter, mark=*, only marks, 
    scatter src=explicit symbolic,
    nodes near coords*={\Label},
    visualization depends on={value \thisrow{label} \as \Label} 
    ] table [meta=class] {
    x y class label
    16 30.17 b MPRNet
    9 31.20 b AirNet
    36 32.06 b PromptIR
    15 28.34 c IDR
    9 25.49 c AirNet
    1 25.09 c TAPE
    };

    \addplot[
    nodes near coords style = {font=\scriptsize, anchor=south west, xshift=-2pt, color=tabblue},
    scatter/classes={
            a={color=tabblue, mark size=1.7},
            d={color=taborange, mark size=1.7}
            },
    scatter, mark=*, only marks, scatter src=explicit symbolic, nodes near coords*={\Label},
    visualization depends on={value \thisrow{label} \as \Label} 
    ] table [meta=class] {
    x y class label
    6 32.51 a Ours 
    };

    \addplot[
    nodes near coords style = {font=\scriptsize, anchor=north west, xshift=-1pt, color=taborange},
    scatter/classes={
            d={color=taborange, mark size=1.7}
            },
    scatter, mark=*, only marks, scatter src=explicit symbolic, nodes near coords*={\Label},
    visualization depends on={value \thisrow{label} \as \Label} 
    ] table [meta=class] {
    x y class label
    6 30.24 d Ours
    };

    \draw[gray, line width=0.75pt] (axis cs:15, 28.34) -- (axis cs:6, 28.34);
    \draw[-stealth, gray, line width=0.75pt] (axis cs:6, 28.34) -- (axis cs:6, 30.04);
    \node at (axis cs:10, 28) {\tiny\textcolor{red}{-$60\%$}};
    \node at (axis cs:1, 29) {\tiny\textcolor{red}{+$1.9$dB}};
    
    \draw[gray, line width=0.75pt] (axis cs:9, 31.20) -- (axis cs:6, 31.20);
    \draw[-stealth, gray, line width=0.75pt] (axis cs:6, 31.20) -- (axis cs:6, 32.25);
    \node at (axis cs:4, 30.95) {\tiny\textcolor{red}{-$33\%$}};
    \node at (axis cs:1, 32) {\tiny\textcolor{red}{+$1.2$dB}};
    
    \draw[draw=tabblue, fill=tabblue] (340, 100) circle [radius=0.05cm];
    \node[text width=1.5cm] at (435,100) {\tiny 3 Degredations};
    \draw[draw=taborange, fill=taborange] (340, 10) circle [radius=0.05cm];
    \node[text width=1.5cm] at (435, 10) {\tiny 5 Degredations};
\end{axis}
\end{tikzpicture}
        \caption{\textit{Model complexity.} Our proposed DaAIR surpasses prior methods, achieving state-of-the-art results in All-in-One image restoration with enhanced efficiency.}
        \label{fig:model_complexity}
    \end{minipage}
\end{figure}

In this paper, we present DaAIR, which provides novel insights for handling multiple degradations concurrently. Specifically, DaAIR employs a dedicated Degradation-aware Learner (DaLe) to learn both the shared and distinct characteristics of each degradation. A novel degradation-aware routing mechanism is further proposed to explicitly associate model capacity with the target degradation, while simultaneously handling agnostic information across degradation types. As illustrated in \cref{fig:motivation}, such a design assigns specialized and shared expertise to address the nuanced aspects of different degradations, unifying task specificity and task agnosticity within a single model without any sort of complex learning, thus setting us apart from previous approaches. We leverage this unification to optimize collaborative learning by projecting specialized and shared features into a low-rank latent space, emphasizing the most informative aspects of both features.

In addition, to further tackle the efficiency aspect of our model, 
we propose a novel cost-efficient parameter update by leveraging the accumulated degradation characteristics from the encoder as an additional supervisory signal for the decoding stage. This self-learnable control mechanism enables cross-task semantic mining, while being guided by the shared common embedding, resulting in better restoration quality. We seamlessly incorporate relevant restoration information with negligible computational overhead. 
This enhances degradation awareness and results in a lightweight, efficient model. Consequently, our design sets a new benchmark for All-in-One image restoration in terms of efficiency and fidelity, as demonstrated in \cref{fig:model_complexity}.

Overall, our key contributions are threefold:
\begin{itemize}
    \item We propose DaAIR which sets a new \textit{state-of-the-art} performance for all-in-one image restoration while being significantly smaller and more efficient compared to prior works.
    \item We propose a dynamic path uniquely associating degradations to explicit experts while harnessing the shared commonalities between degradations to enrich the degradation-awareness of our model. 
    \item  A self-learning mechanism is proposed to utilize the model's inherent information for improving the image quality, achieving seamless integration of task-relevant cues with minimal computational cost.
\end{itemize}

\section{Related Works}

\paragraph{Task Specific Image Restoration.}
Reconstructing the clean image from its degraded counterpart is a highly ill-posed problem, however, a great body of work have addressed image restoration from a data-driven learning perspective, achieving tremendous results compared to prior hand-crafted methods~\cite{zhang2017learning,tai2017memnet,lehtinen2018noise2noise,zhang2019residual,liang2021swinir,Zamir2021Restormer,wang2022uformer}. Most proposed solutions build on convolutional~\cite{zhang2017learning,tai2017memnet,zhang2019residual,chen2022simple} or Transformer-based architectures~\cite{liu2021swin,wang2022uformer,Zamir2021Restormer,chen2022cross} addressing single degradation tasks, such as denoising~\cite{zhang2017learning,zhang2019residual,chen2022cross}, dehazing~\cite{ren2020singledehazing,ren2018gated,wu2021contrastivedehazing} or deraining~\cite{jiang2020multi,ren2019progressive}. Contrary to CNN-based networks, Transformer offer strong modeling capabilities for capturing global dependencies, which makes them outstanding image restorers~\cite{liu2021swin,Zamir2021Restormer,zhao2023comprehensive}. Self-attention's quadratic complexity w.r.t the image size poses a challenge for resource-constrained applications. Conversely, convolutions provide fast and efficient processing, albeit with limited global processing, and scale more effectively with the input size. In this work, we enhance a Transformer-based architecture~\cite{Zamir2021Restormer} by incorporating a dedicated degradation-aware learner for efficiently capturing the shared and distinct context of each degradation. 

\paragraph{All-in-One Image Restoration.}
Research on restoring degraded images has been thorough, but practical implementation is hindered by the need for different models and the challenge of selecting the appropriate one for each task. Images commonly exhibit multiple issues like noise and blur, compounding the difficulty of addressing them individually. An emerging field known as All-in-One image restoration is advancing in low-level computer vision, utilizing a single deep blind restoration model to tackle multiple degradation types simultaneously~\cite{zamir2021pmrnet,valanarasu2022transweather,chen2023alwayscleardays,jiang2023autodir,potlapalli2023promptir,zhang2023ingredient}.
The seminal work, AirNet~\cite{li2022airnet}, achieves blind All-in-One image restoration by employing contrastive learning to extract degradation representations from corrupted images, which are subsequently utilized to restore the clean image.
Next, IDR~\cite{zhang2023ingredient} decomposes image degradations into their underlying physical principles, achieving All-in-One image restoration through a two-stage process based on meta-learning. 
Prompt-based learning~\cite{potlapalli2023promptir, wang2023promptrestorer, li2023pip} has emerged as a promising research direction. Notably, ~\cite{potlapalli2023promptir} introduces tunable prompts that encode discriminative information about degradation types, albeit involving a large number of parameters. In contrast, we propose a novel, parameter-efficient approach that meticulously emphasizes the most informative aspects of the relevant features across various degradations.
\section{Methodology}

\begin{figure}[t]
    \centering
    \includegraphics[width=\textwidth]{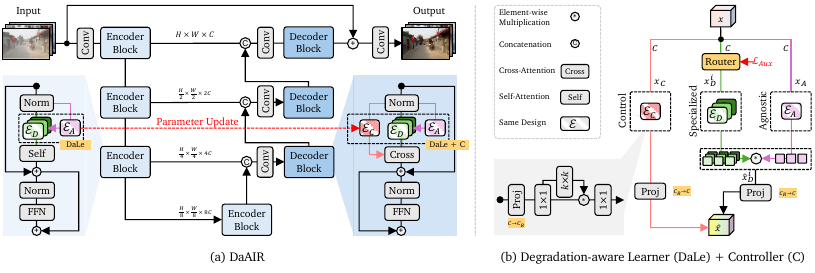}
    \caption{\textit{Architecture overview.} DaAIR reconstructs missing information using an asymmetric encoder-decoder architecture. Each encoder block integrates our proposed Degradation-aware Learner (DaLe) adaptively routing model capacity. Additionally, the decoder blocks are complemented by a controller (C) to enhance overall reconstruction.}
    \label{fig:method:main}
\end{figure}
In this section, we outline the core principles of our efficient All-in-One image restoration method. As shown in \cref{fig:method:main}, our pipeline uses a U-shaped architecture~\cite{ronneberger2015unet,Zamir2021Restormer} with an asymmetric encoder-decoder design. A $3\times3$ convolution first extracts shallow features from the degraded input, which then pass through $4$ levels of encoding and decoding stages. Each level includes several Transformer blocks~\cite{Zamir2021Restormer,potlapalli2023promptir}, incorporating our \textbf{De}gradation-\textbf{a}ware \textbf{L}earner (DaLe) before the attention layer. Unlike previous Transformer-based methods~\cite{liang2021swinir,Zamir2021Restormer,potlapalli2023promptir}, our decoder blocks receive signals from the encoder blocks to help recover the clean output via \textit{controllers}. In each decoder block, we replace the self-attention layer with cross-attention to facilitate interaction between features computed by the DaLe and the controller blocks. Finally, a global residual connection links the shallow features to the output of the refinement stage, capturing high-frequency details before producing the restored image.

\subsection{\textbf{D}egradation-aware \textbf{L}earner}
Our objective is to minimize the overall computational complexity while proficiently encapsulating the spatial data crucial for image restoration.
 We employ efficient experts to modulate the \textit{degradation-specific} and \textit{degradation-agnostic} information within a low-rank framework~\cite{hu2022lora}. Our \textbf{D}egradation-aware \textbf{L}earner (DaLe) consists of an agnostic expert ($\mathcal{E}_{A}$) and specialized experts  ($\mathcal{E}_{D}$) for each degradation type. These experts share a common design, projecting into a low-rank space followed by element-wise multiplication of specialized and agnostic features. This approach enhances degradation-dependent representations with contextual cues, exploring inter-dependencies between various degradations like noise or haze. 
  In the decoder, we integrate a controller block within DaLe to guide the restoration process, leveraging accumulated task-specific knowledge from the optimization phase.

\paragraph{Routing.}
Drawing inspiration from sparse Mixture-of-Experts concepts~\cite{shazeer2017moe,riquelme2021visionmoe}, we integrate linear layers within each DaLe to enable a routing mechanism, associating input features $\mathbf{x} \in \mathbb{R}^{H \times W \times C}$ with their respective specialised degradation experts $\mathcal{E}_{D}$. This routing layer $\mathcal{R}$ orchestrates the allocation of same-task training samples to degradation-specific experts $\mathcal{E}_{D}^{i}$, indexed by $i \in {1,..., N}$, with $N$ representing the total considered degradations, effectively enabling top-$1$ sparse routing. During training, we employ an auxiliary classification loss $\mathcal{L}_{Aux}$ to infuse task-related knowledge into the routing layers, ensuring they explicitly correlate with the target degradation. Without this, the routers lack a basis for selecting specific experts, leading to randomized routing and thus suboptimal model capacity utilization. The impact of the auxiliary loss on the routing is illustrated in \cref{fig:exp:routing}.

\paragraph{Expert design.}  
Each expert contains a projection of their input features into the low-rank approximation $\mathbf{x}_{D} \in \mathbb{R}^{H \times W \times C_{R}}$ with $C_{R} << C$ for emphasizing the most relevant information along the channel dimension, reducing redundancy while allowing for efficient processing in low dimensional space. Next, a large-kernel convolutional block~\cite{hou2022conv2former}, denoted as Conv2Former, captures correlations among pixels within a local neighbourhood, mimicking window-based SA layers~\cite{liang2021swinir,chen2023dat,chen2022cross}, while preserving the efficiency benefits of convolutions. Efficient design for the degradation-specific experts, $\mathcal{E}_{D}$ is more crucial, given the escalating computational demands that arise in direct proportion to the number of experts, which, in turn, is contingent upon the type of degradation under consideration. 
In light of this, as illustrated in \cref{fig:method:main}, the normalized input features $\mathbf{x}$ are dispatched to their associated degradation experts via the router $\mathcal{R}$, partitioning the input batch into groups containing identical degradation training samples. This division allows each expert to specialize within its specific task generating the degradation-specific representation $\mathbf{x}_{D}$. Concurrently, the agnostic expert $\mathcal{E}_{A}$ processes the entire batch, learning a shared representation $\mathbf{x}_{A}$ across all degradations to complement $\mathbf{x}_{D}$ with inter-dependencies among different degradations.
More concretely, the modulated degradation features $\mathbf{\hat{x}}_{D}^{i}$ are obtained as following:
\begin{align}
    \mathbf{\hat{x}}_{D}^{i} &= \mathcal{E}_{D}^{i}(\mathbf{x}_{D}^{i}) \odot \mathcal{E}_{A}(\mathbf{x}_{A})\\
    \text{with}~ \mathcal{E}_{D}^{i}(\mathbf{x}_{D}^{i}) &= \text{Conv2Former}(\mathbf{W}_{C \rightarrow C_{R}}^{1}\mathbf{x}_{D}^{i}) \\
    \text{and}~ \mathcal{E}_{A}(\mathbf{x}_{A})&= \text{Conv2Former}(\mathbf{W}_{C \rightarrow C_{R}}^{2} \mathbf{x}_{A})
\end{align}
where linear layers for compressing the channel dimensions are denoted as $\mathbf{W}$ and $\odot$ denotes the Hadamard product. Next, the sub-batches containing $\mathbf{\hat{x}}_{D}^{i}$ are merged before a shared linear layer expands the channel dimension back to its original size producing the final degradation-aware features $\mathbf{\hat{x}}$.

\subsection{Self-learnable Control}
Prompt-based techniques~\cite{potlapalli2023promptir,li2023pip,wang2023promptrestorer} embed task-specific details into trainable parameters that interact with features to enhance them based on degradation types. However, this approach increases model size and computational demands during both training and inference. Moreover, the knowledge encapsulated in the learned prompt parameters remains opaque, complicating efforts to interpret and improve restoration quality.
To address this challenge without compromising the accuracy-efficiency trade-off, we utilize the knowledge stored in the encoder's agnostic expert to guide parameter updates in the decoder's controller block.
This strategy ensures the optimization process incorporates pertinent information crucial to the restoration task, addressing interpretability concerns linked to prompt-based learning with minimal computational load. 
The controller block shares the same design as $\mathcal{E}_{A}$, allowing to update the learned parameters based on an exponential moving average (EMA) with $\alpha$ as a balancing factor. For guiding the reconstruction process within the decoder branch, we use cross-attention between the controlling features $\mathbf{x}_{C}$ and the modulated features $\mathbf{\hat{x}}$. 
The controller features $\mathbf{x}_{C}$ generate keys, emphasizing pertinent information, while queries and values are derived from modulated features $\mathbf{\hat{x}}$. Cross-attention then refines the restoration process by enabling the model to selectively focus on regions of $\mathbf{\hat{x}}$.
In \cref{tab:exp:high_level_ablations}, we highlight the advantages of our control mechanism. Additionally, the feature visualizations illustrated in \cref{fig:exp:feature_vis} demonstrate the controller's ability to capture degradation characteristics.
\section{Experiments}

We conduct experiments by strictly following previous works in general image restoration~\cite{potlapalli2023promptir,zhang2023ingredient} under two different settings: \textit{(i) All-in-One} and \textit{(ii) Single-task}. In the All-in-One setting, a unified model is trained across multiple degradation types, where we consider \textit{three} and \textit{five} distinct degradations. Within the Single-task setting, separate models are trained for each specific restoration task.

\paragraph{Implementation Details.}
Our DaAIR framework is end-to-end trainable requiring no multi-stage optimization of any components. The architecture comprises a four-level encoder-decoder structure, with each level containing a different number of transformer blocks, namely $[2, 3, 3, 4]$ from highest to lowest level. The reduction ratio between the embedding and low-rank dimensionality for all experts is $16$ and kept constant throughout the network. We follow the training configuration of prior work~\cite{potlapalli2023promptir}, and train our models for $120$ epochs with a batch size of $32$ for All-in-One setting and a batch size of $8$ for the single task setting. We optimize the $L_{1}$ loss using the Adam~\cite{kingma2017adam} optimizer ($\beta_1=0.9$, $\beta_2=0.999$) with an initial learning rate of $\num{2e-4}$ and cosine decay. During training, we utilize crops of size $128^2$ with horizontal and vertical flips as augmentations.

\paragraph{Datasets.}
For All-in-One and single-task settings, we follow existing work~\cite{li2022airnet,potlapalli2023promptir} and include following datasets: For image denoising in single task setting, we combine the BSD400 \cite{arbelaez2010contour} and WED \cite{ma2016waterloo} datasets, adding Gaussian noise at levels $\sigma \in [15, 25, 50]$ to create noisy images. Testing is conducted on the BSD68 \cite{martin2001database} and Urban100 \cite{huang2015single} datasets. For single-task deraining, we use Rain100L \cite{yang2020learning}. The single task dehazing task utilises the SOTS \cite{li2018benchmarking} dataset. For deblurring and low-light enhancement, we employ the GoPro \cite{nah2017deep} and the LOL-v1 \cite{wei2018deep} dataset, respectively. To develop a unified model for all tasks, we merge these datasets in a \textit{three} or \textit{five} degradation setting, and train for 120 epochs and directly evaluate them across different tasks. While, for a single task, our method is trained for 120 epochs on the respective training set.
\begin{table*}[b]
    \centering
    \footnotesize
    \fboxsep0.75pt
    \setlength\tabcolsep{3pt}
    \caption{\textit{Comparison to state-of-the-art on three degradations.} PSNR (dB, $\uparrow$) and \colorbox{clblue!50}{SSIM ($\uparrow$)} metrics are reported on the full RGB images. \textcolor{tabred}{\textbf{Best}} and \textcolor{tabblue}{\textbf{second best}} performances are highlighted. Our method sets a new state-of-the-art on average across all benchmarks while being significantly more efficient than prior work. ‘-’ represents unreported results.}
    \label{tab:exp:3deg}
    \begin{tabularx}{\textwidth}{X*{15}{c}}
    \toprule
     \multirow{2}{*}{Method} & \multirow{2}{*}{Params.} 
     & \multicolumn{2}{c}{\textit{Dehazing}} & \multicolumn{2}{c}{\textit{Deraining}} & \multicolumn{6}{c}{\textit{Denoising}} 
     & \multicolumn{2}{c}{\multirow{2}{*}{Average}}  \\
     \cmidrule(lr){3-4} \cmidrule(lr){5-6} \cmidrule(lr){7-12} 
     && \multicolumn{2}{c}{SOTS} & \multicolumn{2}{c}{Rain100L} & \multicolumn{2}{c}{BSD68\textsubscript{$\sigma$=15}} & \multicolumn{2}{c}{BSD68\textsubscript{$\sigma$=25}} & \multicolumn{2}{c}{BSD68\textsubscript{$\sigma$=50}} &  \\
     \midrule
        BRDNet~\cite{tian2000brdnet} & - & 23.23 &  \cellcolor{clblue!50}{.895} & 27.42 &  \cellcolor{clblue!50}{.895} & 32.26 &  \cellcolor{clblue!50}{.898} & 29.76 &  \cellcolor{clblue!50}{.836} & 26.34 &  \cellcolor{clblue!50}{.693} & 27.80 &  \cellcolor{clblue!50}{.843} \\
        LPNet~\cite{gao2019dynamic} & - & 20.84 &  \cellcolor{clblue!50}{828} & 24.88 &  \cellcolor{clblue!50}{.784} & 26.47 &  \cellcolor{clblue!50}{.778} & 24.77 &  \cellcolor{clblue!50}{.748} & 21.26 &  \cellcolor{clblue!50}{.552} & 23.64 &  \cellcolor{clblue!50}{.738} \\
        FDGAN~\cite{dong2020fdgan}  & - & 24.71 &  \cellcolor{clblue!50}{.929} & 29.89 &  \cellcolor{clblue!50}{.933} & 30.25 &  \cellcolor{clblue!50}{.910} & 28.81 &  \cellcolor{clblue!50}{.868} & 26.43 &  \cellcolor{clblue!50}{.776} & 28.02 &  \cellcolor{clblue!50}{.883} \\
        DL~\cite{fan2019dl} & 2M & 26.92 & \cellcolor{clblue!50}{.931} & 32.62 & \cellcolor{clblue!50}{.931} & 33.05 & \cellcolor{clblue!50}{.914} & 30.41 & \cellcolor{clblue!50}{.861} & 26.90 & \cellcolor{clblue!50}{.740}  & 29.98 & \cellcolor{clblue!50}{.876}\\
        MPRNet~\cite{zamir2021pmrnet}  & 16M & 25.28 &  \cellcolor{clblue!50}{.955} & 33.57 &  \cellcolor{clblue!50}{.954} & 33.54 &  \cellcolor{clblue!50}{.927} & 30.89 &  \cellcolor{clblue!50}{.880} & 27.56 &  \cellcolor{clblue!50}{.779} & 30.17 &  \cellcolor{clblue!50}{.899} \\
        AirNet~\cite{li2022airnet} & 9M & 27.94 &  \cellcolor{clblue!50}{.962} & 34.90 &  \cellcolor{clblue!50}{.967} & 33.92 & \textcolor{tabblue}{\textbf{\cellcolor{clblue!50}{.933}}} & 31.26 &  \textcolor{tabblue}{\textbf{\cellcolor{clblue!50}{.888}}} & 28.00 &  \textcolor{tabblue}{\textbf{\cellcolor{clblue!50}{.797}}} & 31.20 &  \cellcolor{clblue!50}{.910} \\
        PromptIR~\cite{potlapalli2023promptir} & 36M & \textcolor{tabblue}{\textbf{30.58}} &  \textcolor{tabblue}{\textbf{\cellcolor{clblue!50}{.974}}} & \textcolor{tabblue}{\textbf{36.37}} &  \textcolor{tabblue}{\textbf{\cellcolor{clblue!50}{.972}}} & \textcolor{tabred}{\textbf{33.98}} &  \textcolor{tabred}{\textbf{\cellcolor{clblue!50}{.933}}} & \textcolor{tabred}{\textbf{31.31}} &  \textcolor{tabred}{\textbf{\cellcolor{clblue!50}{.888}}} & \textcolor{tabred}{\textbf{28.06}} &  \textcolor{tabred}{\textbf{\cellcolor{clblue!50}{.799}}} & \textcolor{tabblue}{\textbf{32.06}} &  \textcolor{tabblue}{\textbf{\cellcolor{clblue!50}{.913}}} \\
        \midrule
        DaAIR (\textit{ours}) & \textcolor{tabred}{\textbf{6M}} & \textcolor{tabred}{\textbf{32.30}} &  \textcolor{tabred}{\textbf{\cellcolor{clblue!50}{.981}}} & \textcolor{tabred}{\textbf{37.10}} &  \textcolor{tabred}{\textbf{\cellcolor{clblue!50}{.978}}} & \textcolor{tabblue}{\textbf{33.92}} &  \cellcolor{clblue!50}{.930} & \textcolor{tabblue}{\textbf{31.26}} &  \cellcolor{clblue!50}{.884} & \textcolor{tabblue}{\textbf{28.00}} &  \cellcolor{clblue!50}{.792} & \textcolor{tabred}{\textbf{32.51}} &  \textcolor{tabred}{\textbf{\cellcolor{clblue!50}{.913}}} \\

     \bottomrule
        
    \end{tabularx}
    
\end{table*}

\begin{table*}[t]
    \centering
    \footnotesize
    \fboxsep0.75pt
    \setlength\tabcolsep{2.75pt}
    \caption{\textit{Comparison to state-of-the-art on five degradations.} PSNR (dB, $\uparrow$) and \colorbox{clblue!50}{SSIM ($\uparrow$)} metrics are reported on the full RGB images with $(^\ast)$ denoting general image restorers, others are specialized all-in-one approaches. \textcolor{tabred}{\textbf{Best}} and \textcolor{tabblue}{\textbf{second best}} performances are highlighted.}
    \label{tab:exp:5deg}
    \begin{tabularx}{\textwidth}{X*{15}{c}}
    \toprule
     \multirow{2}{*}{Method} & \multirow{2}{*}{Params.} 
     & \multicolumn{2}{c}{\textit{Dehazing}} & \multicolumn{2}{c}{\textit{Deraining}} & \multicolumn{2}{c}{\textit{Denoising}} 
     & \multicolumn{2}{c}{\textit{Deblurring}} & \multicolumn{2}{c}{\textit{Low-Light}} & \multicolumn{2}{c}{\multirow{2}{*}{Average}}  \\
     \cmidrule(lr){3-4} \cmidrule(lr){5-6} \cmidrule(lr){7-8} \cmidrule(lr){9-10} \cmidrule(lr){11-12}
     && \multicolumn{2}{c}{SOTS} & \multicolumn{2}{c}{Rain100L} & \multicolumn{2}{c}{BSD68\textsubscript{$\sigma$=25}} 
     & \multicolumn{2}{c}{GoPro} & \multicolumn{2}{c}{LOLv1} &  \\
     \midrule
     
    NAFNet$^\ast$~\cite{chen2022simple} & 17M & 25.23 & \cellcolor{clblue!50}{.939} & 35.56 & \cellcolor{clblue!50}{.967} & 31.02 &  \cellcolor{clblue!50}{.883} & 26.53 &  \cellcolor{clblue!50}{.808} & 20.49 &  \cellcolor{clblue!50}{.809} & 27.76 &  \cellcolor{clblue!50}{.881} \\
    DGUNet$^\ast$~\cite{mou2022dgunet} & 17M & 24.78 & \cellcolor{clblue!50}{.940} & 36.62 & \cellcolor{clblue!50}{.971} & 31.10 &  \cellcolor{clblue!50}{.883} & 27.25 &  \cellcolor{clblue!50}{.837} & 21.87 &  \cellcolor{clblue!50}{.823} & 28.32 &  \cellcolor{clblue!50}{.891} \\
    SwinIR$^\ast$~\cite{liang2021swinir} & 1M & 21.50 & \cellcolor{clblue!50}{.891} & 30.78 & \cellcolor{clblue!50}{.923} & 30.59 &  \cellcolor{clblue!50}{.868} & 24.52 &  \cellcolor{clblue!50}{.773} & 17.81 &  \cellcolor{clblue!50}{.723} & 25.04 &  \cellcolor{clblue!50}{.835} \\ 
    Restormer$^\ast$~\cite{Zamir2021Restormer} & 26M & 24.09 & \cellcolor{clblue!50}{.927} & 34.81 & \cellcolor{clblue!50}{.962} & 31.49 &  \cellcolor{clblue!50}{.884} & 27.22 &  \cellcolor{clblue!50}{.829} & 20.41 &  \cellcolor{clblue!50}{.806} & 27.60 &  \cellcolor{clblue!50}{.881} \\
    \midrule
    
    DL~\cite{fan2019dl} & 2M & 20.54 &  \cellcolor{clblue!50}{.826} & 21.96 &  \cellcolor{clblue!50}{.762} & 23.09 &  \cellcolor{clblue!50}{.745} & 19.86 &  \cellcolor{clblue!50}{.672} & 19.83 &  \cellcolor{clblue!50}{.712} & 21.05 &  \cellcolor{clblue!50}{.743} \\
    Transweather~\cite{valanarasu2022transweather} & 38M & 21.32 &  \cellcolor{clblue!50}{.885} & 29.43 &  \cellcolor{clblue!50}{.905} & 29.00 &  \cellcolor{clblue!50}{.841} & 25.12 &  \cellcolor{clblue!50}{.757} & 21.21 &  \cellcolor{clblue!50}{.792} & 25.22 &  \cellcolor{clblue!50}{.836} \\
    TAPE~\cite{liu2022tape} & 1M & 22.16 &  \cellcolor{clblue!50}{.861} & 29.67 &  \cellcolor{clblue!50}{.904} & 30.18 &  \cellcolor{clblue!50}{.855} & 24.47 &  \cellcolor{clblue!50}{.763} & 18.97 &  \cellcolor{clblue!50}{.621} & 25.09 &  \cellcolor{clblue!50}{.801} \\
    AirNet~\cite{li2022airnet} & 9M & 21.04 &  \cellcolor{clblue!50}{.884} & 32.98 &  \cellcolor{clblue!50}{.951} & 30.91 &  \textcolor{tabblue}{\textbf{\cellcolor{clblue!50}{.882}}} & 24.35 &  \cellcolor{clblue!50}{.781} & 18.18 &  \cellcolor{clblue!50}{.735} & 25.49 &  \cellcolor{clblue!50}{.847} \\
      IDR~\cite{zhang2023ingredient} & 15M & \textcolor{tabblue}{\textbf{25.24}} &  \textcolor{tabblue}{\textbf{\cellcolor{clblue!50}{.943}}} & \textcolor{tabblue}{\textbf{35.63}} &  \textcolor{tabblue}{\textbf{\cellcolor{clblue!50}{.965}}} & \textcolor{tabred}{\textbf{31.60}} &  \textcolor{tabred}{\textbf{\cellcolor{clblue!50}{.887}}} & \textcolor{tabblue}{\textbf{27.87}} &  \textcolor{tabblue}{\textbf{\cellcolor{clblue!50}{.846}}} & \textcolor{tabblue}{\textbf{21.34}} &  \textcolor{tabred}{\textbf{\cellcolor{clblue!50}{.826}}} & \textcolor{tabblue}{\textbf{28.34}} &  \textcolor{tabblue}{\textbf{\cellcolor{clblue!50}{.893}}} \\
      
     \midrule
     DaAIR (\textit{ours}) &  \textcolor{tabred}{\textbf{6M}} &  \textcolor{tabred}{\textbf{31.97}} &   \textcolor{tabred}{\textbf{\cellcolor{clblue!50}{.980}}} &  \textcolor{tabred}{\textbf{36.28}} &   \textcolor{tabred}{\textbf{\cellcolor{clblue!50}{.975}}} &  \textcolor{tabblue}{\textbf{31.07}} &  \cellcolor{clblue!50}{.878} &  \textcolor{tabred}{\textbf{29.51}} &   \textcolor{tabred}{\textbf{\cellcolor{clblue!50}{.890}}} & \textcolor{tabred}{\textbf{22.38}} &   \textcolor{tabblue}{\textbf{\cellcolor{clblue!50}{.825}}} &  \textcolor{tabred}{\textbf{30.24}} &   \textcolor{tabred}{\textbf{\cellcolor{clblue!50}{.910}}} \\
     \bottomrule
    \end{tabularx}
\end{table*}

\begin{table*}[t]
    \centering
    \footnotesize
    \fboxsep0.75pt  
    \setlength\tabcolsep{1pt}
    \caption{\textit{Comparison to state-of-the-art for single degradations.} PSNR (dB, $\uparrow$) and \colorbox{clblue!50}{SSIM ($\uparrow$)} metrics are reported on the full RGB images. \textcolor{tabred}{\textbf{Best}} and \textcolor{tabblue}{\textbf{second best}} performances are highlighted. Our method excels prior work on dehazing and deraining. 
    }
    \label{tab:exp:single}

    \begin{subtable}[l]{0.25\textwidth}
        \subcaption{\textit{Dehazing}}
        \begin{tabularx}{\textwidth}{X*{3}{c}}
        \toprule
        Method & \multicolumn{2}{c}{SOTS}\\
        \midrule
        DehazeNet ~\cite{cai2016dehazenet} & 22.46 & \cellcolor{clblue!50}{.851}\\
        EPDN ~\cite{qu2019enhanced} & 22.57 & \cellcolor{clblue!50}{.863}\\
        FDGAN~\cite{dong2020fdgan} & 23.15 & \cellcolor{clblue!50}{.921} \\
        \midrule
        AirNet~\cite{li2022airnet} & 23.18 & \cellcolor{clblue!50}{.900} \\
        PromptIR~\cite{potlapalli2023promptir} & \textcolor{tabblue}{\textbf{31.31}} & \textcolor{tabblue}{\textbf{\cellcolor{clblue!50}{.973}}} \\
        \midrule 
        DaAIR (\textit{ours}) & \textcolor{tabred}{\textbf{31.99}} & \textcolor{tabred}{\textbf{.981}} \\
        \bottomrule
        \end{tabularx}
    \end{subtable}%
    \hfill
    \begin{subtable}[l]{0.25\textwidth}
        \subcaption{\textit{Deraining}}
        \begin{tabularx}{\textwidth}{X*{3}{c}}
        \toprule
        Method & \multicolumn{2}{c}{Rain100L}\\
        \midrule
        UMR ~\cite{yasarla2019uncertainty} & 32.39 & \cellcolor{clblue!50}{.921} \\
        MSPFN ~\cite{jiang2020multi} & 33.50 & \cellcolor{clblue!50}{.948} \\
        LPNet~\cite{gao2019dynamic}  & 23.15 & \cellcolor{clblue!50}{.921} \\
        \midrule
        AirNet~\cite{li2022airnet}  & 34.90 & \cellcolor{clblue!50}{.977} \\
        PromptIR~\cite{potlapalli2023promptir} & \textcolor{tabblue}{\textbf{37.04}} & \textcolor{tabblue}{\textbf{\cellcolor{clblue!50}{.979}}} \\
        \midrule 
         DaAIR (\textit{ours}) & \textcolor{tabred}{\textbf{37.78}} & \cellcolor{clblue!50}{\textcolor{tabred}{\textbf{.982}}}\\
        \bottomrule
        \end{tabularx}
    \end{subtable}%
    \hfill
    \begin{subtable}[l]{0.45\textwidth}
        \subcaption{\textit{Denoising}}
        \begin{tabularx}{\textwidth}{X*{6}{c}}
        \toprule
        Method & \multicolumn{2}{c}{$\sigma$=15} & \multicolumn{2}{c}{$\sigma$=25} & \multicolumn{2}{c}{$\sigma$=50}\\
        \midrule
        IRCNN ~\cite{zhang2017learning}  & 33.87 & \cellcolor{clblue!50}{.929} & 31.18 & \cellcolor{clblue!50}{.882} & 27.88 & \cellcolor{clblue!50}{.790}  \\
        FFDNet ~\cite{zhang2018ffdnet} & 33.87 & \cellcolor{clblue!50}{.929} & 31.21 & \cellcolor{clblue!50}{.882} & 27.96 & \cellcolor{clblue!50}{.789} \\
        \midrule 
        BRDNet~\cite{tian2000brdnet} & 34.10 & \cellcolor{clblue!50}{.929} & 31.43 & \cellcolor{clblue!50}{.885} & 28.16 & \cellcolor{clblue!50}{.794} \\ 
        AirNet~\cite{li2022airnet} & 34.14 & \cellcolor{clblue!50}{.936} & 31.48 & \cellcolor{clblue!50}{.893} & 28.23 & \cellcolor{clblue!50}{.806}  \\
        PromptIR~\cite{potlapalli2023promptir} & \textcolor{tabred}{\textbf{34.34}} & \textcolor{tabred}{\textbf{\cellcolor{clblue!50}{.938}}} & \textcolor{tabred}{\textbf{31.71}} & \textcolor{tabred}{\textbf{\cellcolor{clblue!50}{.897}}} & \textcolor{tabred}{\textbf{28.49}} & \textcolor{tabred}{\textbf{\cellcolor{clblue!50}{.813}}} \\
        \midrule 
         DaAIR (\textit{ours})  & \textcolor{tabblue}{\textbf{34.25}} & \textcolor{tabblue}{\textbf{\cellcolor{clblue!50}{.934}}} & \textcolor{tabblue}{\textbf{31.61}} & \textcolor{tabblue}{\textbf{\cellcolor{clblue!50}{.891}}} & \textcolor{tabblue}{\textbf{28.36}} &\textcolor{tabblue}{\textbf{\cellcolor{clblue!50}{.807}}} \\
        \bottomrule
        \end{tabularx}
    \end{subtable}
\end{table*}

\subsection{Comparison to State-of-the-Art Methods}
\paragraph{All-in-One: Three Degradations.}
We compare our All-in-One restorer with specialized All-in-One restoration methods, including BRDNet~\cite{tian2000brdnet}, LPNet~\cite{gao2019dynamic}, FDGAN~\cite{dong2020fdgan}, DL~\cite{fan2019dl}, MPRNet~\cite{zamir2021pmrnet}, AirNet~\cite{li2022airnet}, and PromptIR~\cite{potlapalli2023promptir}, trained simultaneously on three degradations: dehazing, deraining, and denoising. Our proposed method emerges as the best All-in-One restorer and the most efficient, as demonstrated in \cref{tab:exp:3deg}. It consistently outperforms previous works, with an average improvement of $0.45$ dB across all benchmarks. Notably, our method achieves state-of-the-art performance on the SOTS and Rain100L benchmarks, surpassing the previously best PromptIR~\cite{potlapalli2023promptir} by $1.72$ dB and $0.72$ dB, respectively, featuring about $83\%$ lesser parameters and $80\%$ lesser GMACS. 

\paragraph{All-in-One: Five Degradations.}
Following recent studies~\cite{li2022airnet,zhang2023ingredient}, we extend the three degradation settings to include deblurring and low-light image enhancement, validating our method's effectiveness in a more complex All-in-One setting. As shown in \cref{tab:exp:5deg}, our method excels by learning dedicated experts for each degradation while modelling the commonalities between tasks. It outperforms AirNet~\cite{li2022airnet} and IDR~\cite{zhang2023ingredient} by $4.75$ dB and $1.9$ dB on average across all five benchmarks, using $33\%$ and $60\%$ fewer parameters, respectively. Additionally, we compare our method to general image restoration models trained in the same All-in-One setting. Notably, our method surpasses Restormer~\cite{Zamir2021Restormer} and NAFNet~\cite{chen2022simple} on the GoPro test set by $2.29$ dB and $2.98$ dB, respectively, while being four times and three times smaller in size.

\paragraph{Single-Degradation Results.}
To assess the efficacy of our proposed framework, we present results in \cref{tab:exp:single} wherein individual instances of our method are trained using the single degradation protocol. 
The single-task variant trained for dehazing consistently outperforms AirNet~\cite{li2022airnet} and PromptIR~\cite{potlapalli2023promptir} by $8.81$ dB and $0.68$ dB, respectively. Similarly, when trained for image deraining, our method surpasses both previous approaches by $3.07$ dB and $0.74$ dB, respectively, demonstrating notable performance enhancements. Additionally, for denoising at different noise ratios, our method performs on par with the second-best method, AirNet~\cite{li2022airnet} proving the generality of our approach across multiple degradations. 

\begin{figure}[t]
    \centering
    \footnotesize
    \begin{subfigure}{\textwidth}
    \begin{tblr}{
      colspec = {@{}lX[c]X[c]X[c]X[c]X[c]@{}},
      colsep=0.1pt,
      rows={rowsep=0.75pt},
      stretch = 0,
    }
     &  Input & AirNet~\cite{li2022airnet} & PromptIR~\cite{potlapalli2023promptir} & Ours & GT \\
        \SetCell[r=2]{l}{\rotatebox{90}{Dehazing}} &       
        \includegraphics[width=0.19\textwidth, height=0.12\textwidth]{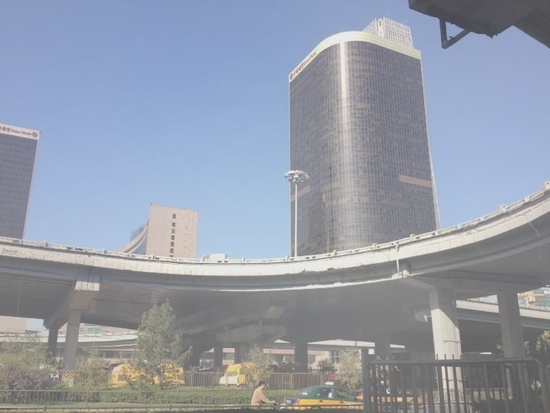} & \includegraphics[width=0.19\textwidth, height=0.12\textwidth, trim={150pt 150pt 25pt 25pt}, clip]{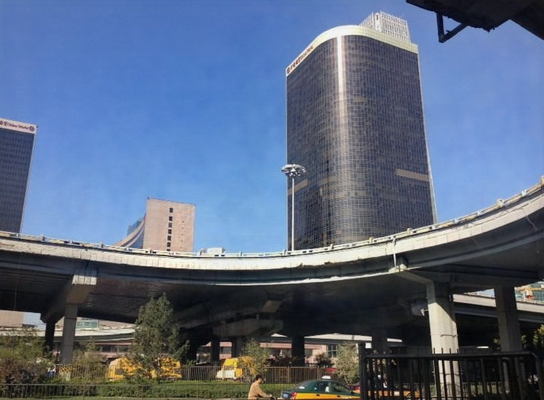} & 
        \includegraphics[width=0.19\textwidth, height=0.12\textwidth, trim={150pt 150pt 25pt 25pt}, clip]{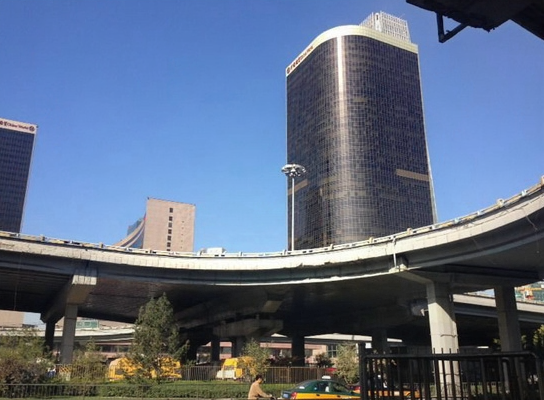} & 
        \includegraphics[width=0.19\textwidth, height=0.12\textwidth, trim={150pt 150pt 25pt 25pt}, clip]{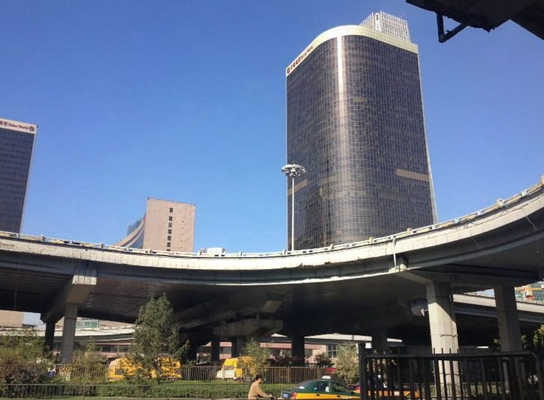} & 
        \includegraphics[width=0.19\textwidth, height=0.12\textwidth, trim={150pt 150pt 25pt 25pt}, clip]{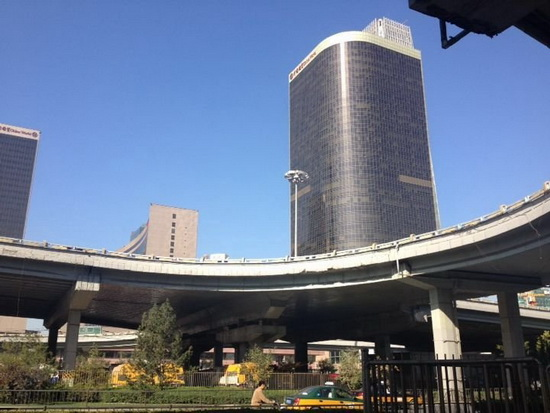}\\

        & \includegraphics[width=0.19\textwidth, height=0.12\textwidth]{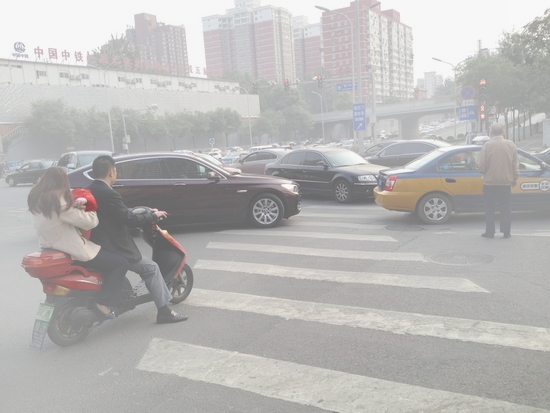} & \includegraphics[width=0.19\textwidth, height=0.12\textwidth, trim={150pt 150pt 80 0}, clip]{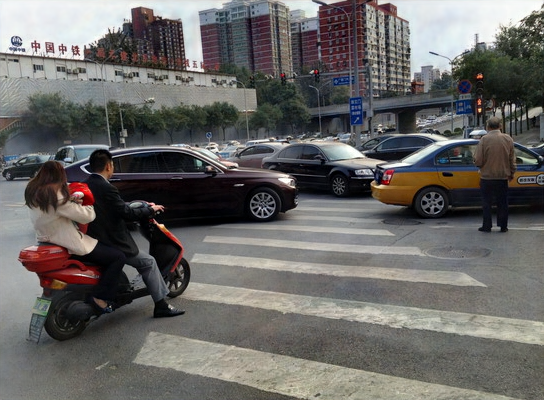} & 
        \includegraphics[width=0.19\textwidth, height=0.12\textwidth, trim={150pt 150pt 80 0}, clip]{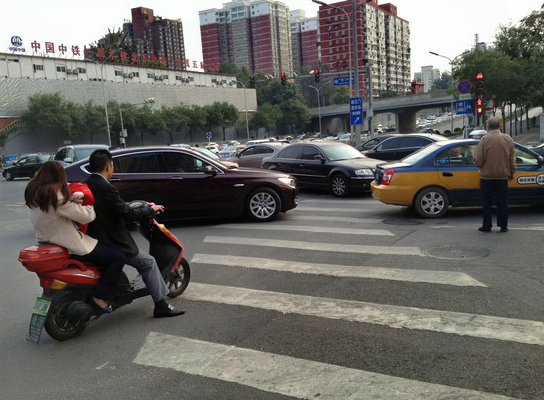} & 
        \includegraphics[width=0.19\textwidth, height=0.12\textwidth, trim={150pt 150pt 80 0}, clip]{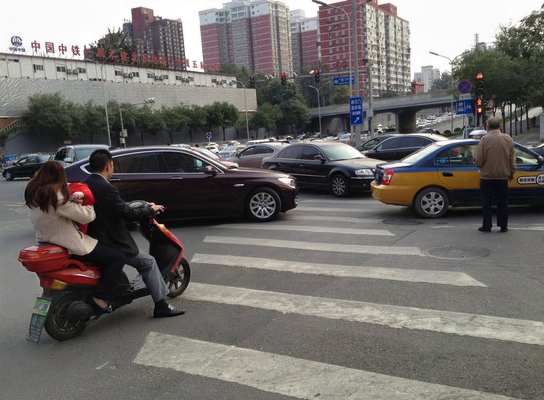} & 
        \includegraphics[width=0.19\textwidth, height=0.12\textwidth, trim={150pt 150pt 80 0}, clip
        ]{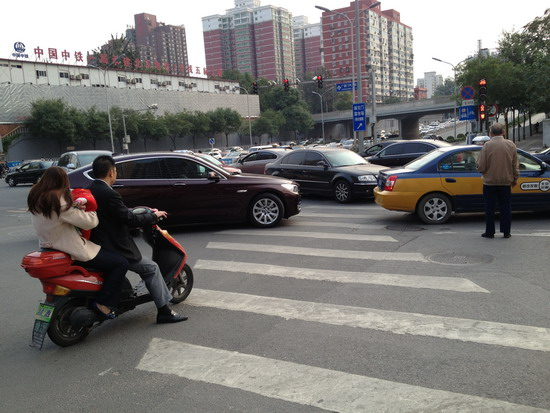}\\
    \end{tblr}
    \end{subfigure}
\hfill
    \begin{subfigure}{\textwidth}
        \begin{tblr}{
          colspec = {@{}lX[c]X[c]X[c]X[c]X[c]@{}},
          colsep=0.1pt,
          rows={rowsep=0.75pt},
          stretch = 0,
    }
        \SetCell[r=2]{l}{\rotatebox{90}{Deraining}} & 
        \includegraphics[width=0.19\textwidth, height=0.12\textwidth, ]{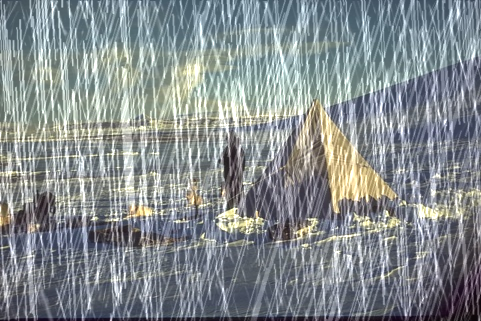} & \includegraphics[width=0.19\textwidth, height=0.12\textwidth, trim={150pt 150pt 80pt 0}, clip]{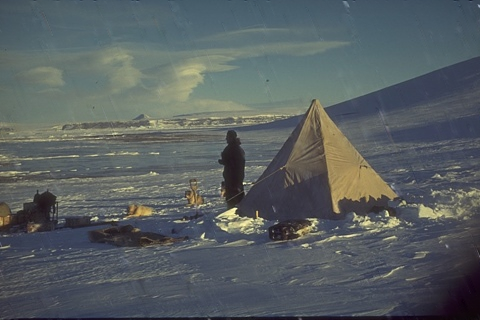} & 
        \includegraphics[width=0.19\textwidth, height=0.12\textwidth, trim={150pt 150pt 80pt 0}, clip]{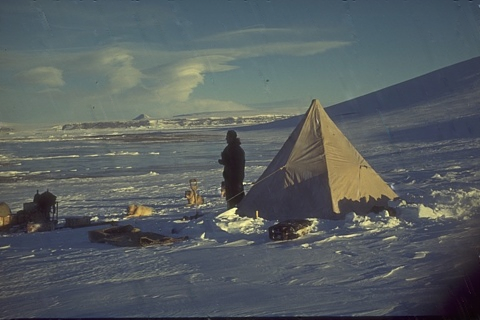} & \includegraphics[width=0.19\textwidth, height=0.12\textwidth, trim={150pt 150pt 80pt 0}, clip]{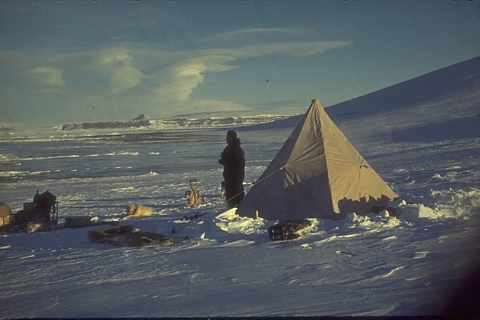} & 
        \includegraphics[width=0.19\textwidth, height=0.12\textwidth, trim={150pt 150pt 80pt 0}, clip]{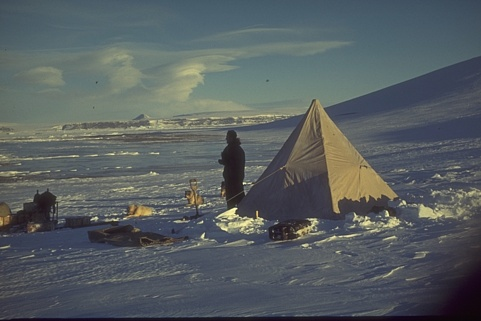}\\
              
        & \includegraphics[width=0.19\textwidth, height=0.12\textwidth]{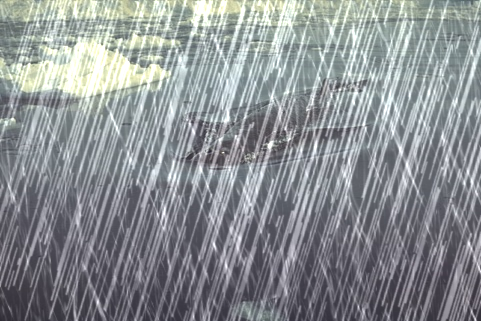} & \includegraphics[width=0.19\textwidth, height=0.12\textwidth, trim={0 0 100pt 100pt}, clip]{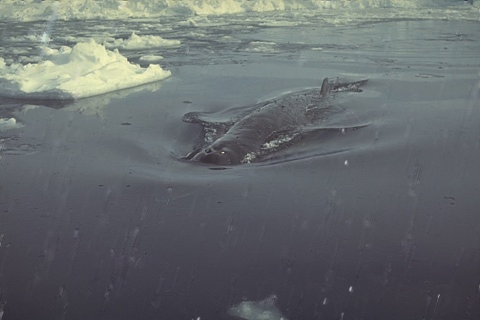} & 
        \includegraphics[width=0.19\textwidth, height=0.12\textwidth, trim={0 0 100pt 100pt}, clip]{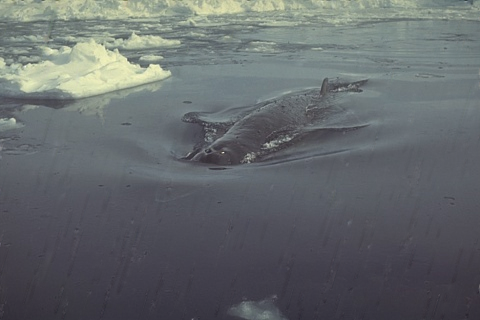} & \includegraphics[width=0.19\textwidth, height=0.12\textwidth, trim={0 0 100pt 100pt}, clip]{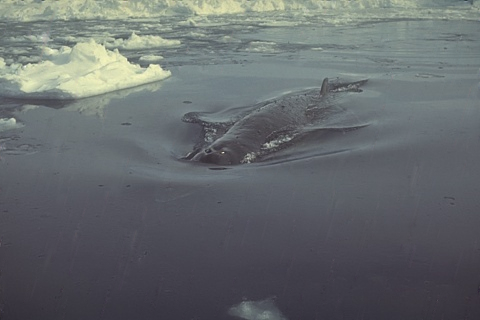} & 
        \includegraphics[width=0.19\textwidth, height=0.12\textwidth, trim={0 0 100pt 100pt}, clip]{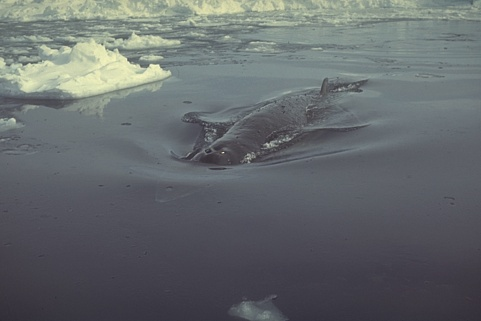}\\
    \end{tblr}
    \end{subfigure}
\hfill
    \begin{subfigure}{\textwidth}
        \begin{tblr}{
          colspec = {@{}lX[c]X[c]X[c]X[c]X[c]@{}},
          colsep=0.1pt,
          rows={rowsep=0.75pt},
          stretch = 0,
        }
            \SetCell[r=2]{l}{\rotatebox{90}{Denoising}} &  
            \includegraphics[width=0.19\textwidth, height=0.12\textwidth]{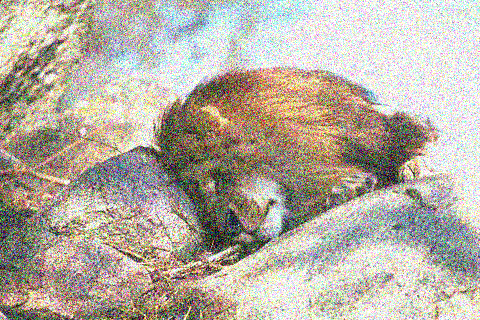} & \includegraphics[width=0.19\textwidth, height=0.12\textwidth, trim={100pt 100pt 100pt 100pt}, clip]{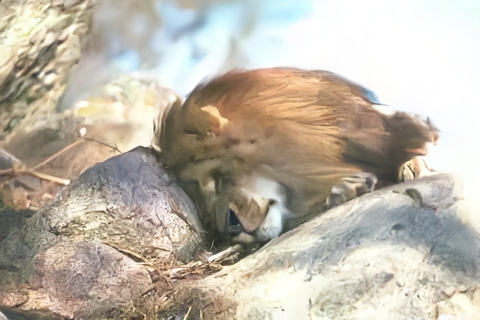} & 
            \includegraphics[width=0.19\textwidth, height=0.12\textwidth, trim={100pt 100pt 100pt 100pt}, clip]{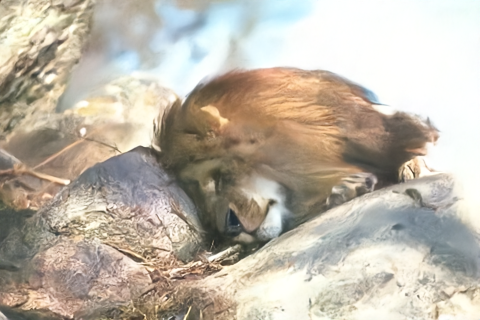} & 
            \includegraphics[width=0.19\textwidth, height=0.12\textwidth, trim={100pt 100pt 100pt 100pt}, clip]{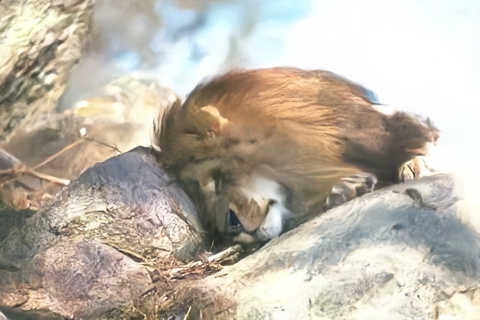} & 
            \includegraphics[width=0.19\textwidth, height=0.12\textwidth, trim={100pt 100pt 100pt 100pt}, clip]{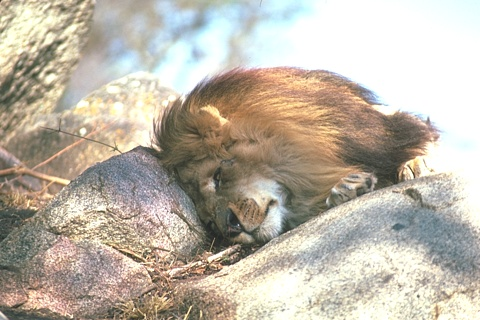}\\
                  
            & \includegraphics[width=0.19\textwidth, height=0.12\textwidth]{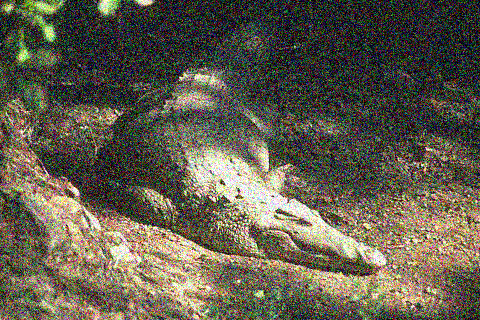} & \includegraphics[width=0.19\textwidth, height=0.12\textwidth, trim={100pt 50pt 100pt 70pt}, clip]{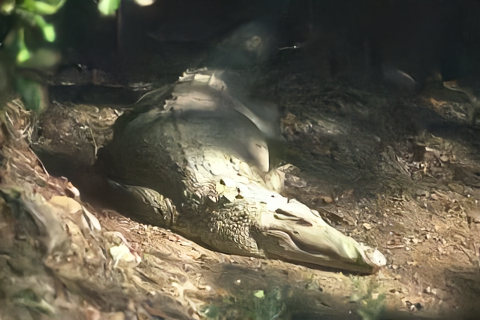} & 
            \includegraphics[width=0.19\textwidth, height=0.12\textwidth, trim={100pt 50pt 100pt 70pt}, clip]{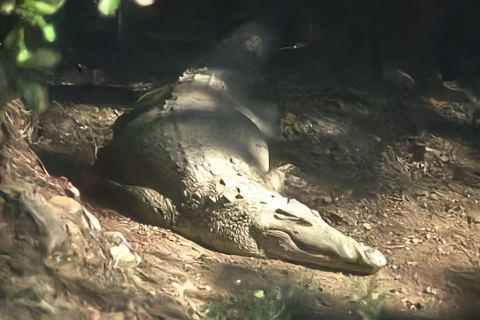} & 
            \includegraphics[width=0.19\textwidth, height=0.12\textwidth, trim={100pt 50pt 100pt 70pt}, clip]{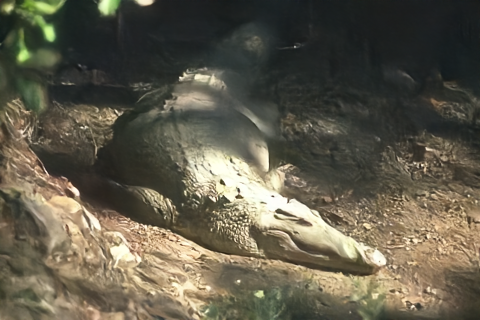} & 
            \includegraphics[width=0.19\textwidth, height=0.12\textwidth, trim={100pt 50pt 100pt 70pt}, clip]{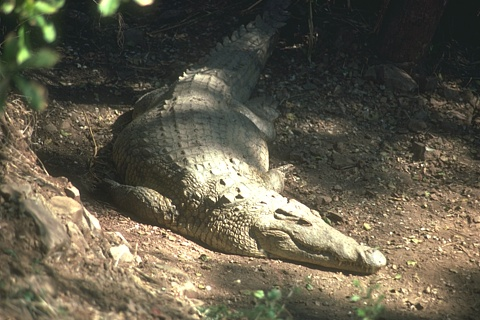}\\
        \end{tblr}
    \end{subfigure}
    \caption{Visual comparison of DaAIR with state-of-the-art methods on challenging cases for the All-in-One setting considering three degradations.}
    \label{fig:exp:visual_results}
\end{figure}

\begin{figure}[t]
    \centering
    \begin{minipage}[t]{0.49\textwidth}
        \centering
        \footnotesize
        \fboxsep0.75pt
        \setlength\tabcolsep{2.5pt}
        \captionof{table}{\textit{Impact of key components.} PSNR (dB, $\uparrow$) and \colorbox{clblue!50}{SSIM ($\uparrow$)} metrics are reported on the full RGB images.}
        \label{tab:exp:high_level_ablations}
        \begin{tabularx}{\textwidth}{X*{6}{c}}
            \toprule
                Method & $\mathcal{E}_{\text{D}}$ & $\mathcal{E}_{\text{A}}$ & $\mathcal{E}_{\text{C}}$ & $\mathcal{L}_{Aux}$ & PSNR & SSIM \\
             \midrule
             Baseline  & - & - & - & - & 31.86 & .911\\
            \textit{(a)} & - & \checkmark  & \checkmark & -  & 32.12 & .911 \\
            \textit{(b)} & \checkmark & - & \checkmark & \checkmark  & 32.41 & .913 \\
            \textit{(c)} & \checkmark & \checkmark &  - & \checkmark  &  32.44	& .913\\
            \textit{(d)} &  \checkmark & \checkmark & \checkmark & - & 32.23 & .912\\
            \midrule
             DaAIR (\textit{ours}) & \checkmark & \checkmark & \checkmark & \checkmark & \textcolor{tabred}{\textbf{32.51}} & \textcolor{tabred}{\textbf{.913}}\\
             \bottomrule
        \end{tabularx}
    \end{minipage}
    \hfill
    \begin{minipage}[t]{0.49\textwidth}
    \centering
    \footnotesize
    \setlength\tabcolsep{2.5pt}
    \captionof{table}{\textit{Complexity Analysis.} GMACS are computed on an input image of size $224\times224$ using a NVIDIA RTX 4090 GPU. 
    }
    \label{tab:exp:efficiency_overview}
        \begin{tabularx}{\textwidth}{X*{3}{c}}
            \toprule
            Method  & Memory & Params. & GMACS \\
            \midrule
            AirNet~\cite{li2022airnet} & 4829M &  8.93M  & 238G \\
            PromptIR~\cite{potlapalli2023promptir} & 9830M & 35.59M & 132G \\
            IDR~\cite{zhang2023ingredient} & 4905M & 15.34M & 98G \\
            \midrule
            DaAIR (\textit{ours})& \textcolor{tabred}{\textbf{3333M}} &  \textcolor{tabred}{\textbf{6.45M}} & \textcolor{tabred}{\textbf{21G}} \\ 
            \bottomrule
        \end{tabularx}
    \end{minipage}
\end{figure}

\begin{figure}[t]
    \centering
    \begin{minipage}[b]{0.49\textwidth}
\begin{tikzpicture}[baseline]
  \begin{groupplot}[
  /pgf/bar width=1.3pt,
  group style={group size=2 by 2, horizontal sep=0.1cm, vertical sep=0.1cm},
  width=0.62\textwidth, height=0.4\textwidth, 
  xtick pos=left, ytick pos=left, xlabel near ticks, ylabel near ticks, xtick=data, ymajorgrids=true,
  grid style=dashed, tick label style={font=\tiny}, title style={font=\tiny}, label style={font=\tiny}, tick align=inside,
  ybar stacked, axis on top, enlargelimits=0.03,
  symbolic x coords={1, 2, 3, 4, 5, 6, 7, 8, 9, 10, 11, 12, 13, 14, 15, 16, 17, 18, 19, 20},
  legend image post style={scale=0.3}, legend style={at={(1.12,-1.1)}, draw=none, fill=none, anchor=south, legend columns=1, font=\tiny,},
  ylabel style={anchor=center, at={(-0.25,0.5)},},
  ]

    \nextgroupplot[
    xticklabels=\empty, ylabel={Deraining}, 
    title={w/o $\mathcal{L}_{Aux}$}, title style={anchor=north, at={(0.5,1.15)},},
    ]
    \addplot[draw=tabblue,fill=tabblue] coordinates {
    (1,1) (2,72) (3,12) (4,15) (5,65) (6,0) (7,4) (8,12) (9,91) (10,0) (11,4) (12,3) (13,1) (14,0) (15,2) (16,10) (17,0) (18,0) (19,0) (20,76)
    }; 
    \addplot[draw=taborange,fill=taborange] coordinates {
    (1,0) (2,1) (3,0) (4,73) (5,28) (6,73) (7,24) (8,0) (9,0) (10,83) (11,16) (12,0) (13,2) (14,45) (15,6) (16,0) (17,17) (18,18) (19,58) (20,4)
    }; 
    \addplot[draw=tabgreen,fill=tabgreen] coordinates {
    (1,3) (2,4) (3,8) (4,8) (5,0) (6,27) (7,56) (8,45) (9,9) (10,1) (11,67) (12,34) (13,94) (14,0) (15,0) (16,12) (17,29) (18,82) (19,25) (20,0)
    }; 
    \addplot[draw=tabred,fill=tabred] coordinates {
    (1,96) (2,17) (3,56) (4,0) (5,0) (6,0) (7,0) (8,32) (9,0) (10,10) (11,13) (12,0) (13,3) (14,47) (15,76) (16,0) (17,16) (18,0) (19,0) (20,13)
    }; 
    \addplot[draw=tabpurple,fill=tabpurple] coordinates {
    (1,0) (2,6) (3,24) (4,4) (5,7) (6,0) (7,16) (8,11) (9,0) (10,6) (11,0) (12,63) (13,0) (14,8) (15,16) (16,78) (17,38) (18,0) (19,17) (20,7)
    }; 

    \nextgroupplot[
    xticklabels=\empty, yticklabels=\empty,
    title={with $\mathcal{L}_{Aux}$}, title style={anchor=north, at={(0.5,1.15)},},
    ]
    \addplot[draw=tabblue,fill=tabblue] coordinates {
    (1,2) (2,0) (3,0) (4,0) (5,0) (6,0) (7,0) (8,0) (9,0) (10,0) (11,0) (12,0) (13,0) (14,0) (15,0) (16,0) (17,0) (18,0) (19,0) (20,0)
    }; 
    \addplot[draw=taborange,fill=taborange] coordinates {
    (1,1) (2,0) (3,0) (4,0) (5,0) (6,0) (7,0) (8,0) (9,0) (10,0) (11,0) (12,0) (13,0) (14,0) (15,0) (16,0) (17,0) (18,0) (19,0) (20,0)
    }; 
    \addplot[draw=tabgreen,fill=tabgreen] coordinates {
    (1,0) (2,0) (3,0) (4,0) (5,0) (6,0) (7,0) (8,0) (9,0) (10,0) (11,0) (12,0) (13,0) (14,0) (15,0) (16,0) (17,0) (18,0) (19,0) (20,0)
    }; 
    \addplot[draw=tabred,fill=tabred] coordinates {
    (1,44) (2,0) (3,0) (4,0) (5,0) (6,0) (7,0) (8,0) (9,0) (10,0) (11,0) (12,0) (13,0) (14,0) (15,0) (16,0) (17,0) (18,0) (19,0) (20,0)
    }; 
    \addplot[draw=tabpurple,fill=tabpurple] coordinates {
    (1,53) (2,100) (3,100) (4,100) (5,100) (6,100) (7,100) (8,100) (9,100) (10,100) (11,100) (12,100) (13,100) (14,100) (15,100) (16,100) (17,100) (18,100) (19,100) (20,100)
    }; 
    
    \legend{$\mathcal{E}_1$, $\mathcal{E}_2$, $\mathcal{E}_3$, $\mathcal{E}_4$, $\mathcal{E}_5$}

    \nextgroupplot[
    xticklabels=\empty, ylabel={Dehazing},
    extra x ticks={3, 11, 18}, extra x tick labels={Encoder, Bottleneck, Decoder},
    xlabel=Layers, every axis x label/.append style={at=(ticklabel cs:1)},
    ]
    \addplot+[draw=tabblue,fill=tabblue,] coordinates {
    (1,0) (2,465) (3,31) (4,178) (5,29) (6,447) (7,234) (8,40) (9,0) (10,0) (11,2) (12,291) (13,44) (14,0) (15,0) (16,56) (17,373) (18,47) (19,380) (20,44)
    }; 
    \addplot[draw=taborange,fill=taborange] coordinates {
    (1,0) (2,0) (3,5) (4,188) (5,230) (6,0) (7,86) (8,0) (9,412) (10,30) (11,0) (12,209) (13,23) (14,82) (15,20) (16,10) (17,27) (18,27) (19,51) (20,30)
    }; 
    \addplot[draw=tabgreen,fill=tabgreen] coordinates {
    (1,0) (2,6) (3,129) (4,10) (5,129) (6,37) (7,2) (8,387) (9,25) (10,38) (11,85) (12,0) (13,0) (14,0) (15,103) (16,45) (17,22) (18,378) (19,27) (20,25)
    }; 
    \addplot[draw=tabred,fill=tabred] coordinates {
    (1,500) (2,20) (3,332) (4,91) (5,55) (6,16) (7,55) (8,68) (9,22) (10,27) (11,0) (12,0) (13,405) (14,47) (15,377) (16,18) (17,25) (18,19) (19,16) (20,375)
    }; 
    \addplot[draw=tabpurple,fill=tabpurple] coordinates {
    (1,0) (2,9) (3,3) (4,33) (5,57) (6,0) (7,123) (8,5) (9,41) (10,405) (11,413) (12,0) (13,28) (14,371) (15,0) (16,371) (17,53) (18,29) (19,26) (20,26)
    }; 

    \nextgroupplot[
    xticklabels=\empty,yticklabels=\empty,
    extra x ticks={3, 11, 18}, extra x tick labels={Encoder, Bottleneck, Decoder},
    ]
    \addplot[draw=tabblue,fill=tabblue] coordinates {
    (1,0) (2,0) (3,0) (4,0) (5,0) (6,0) (7,0) (8,0) (9,0) (10,0) (11,0) (12,0) (13,0) (14,0) (15,0) (16,0) (17,0) (18,0) (19,0) (20,0)
    }; 
    \addplot[draw=taborange,fill=taborange] coordinates {
    (1,0) (2,0) (3,0) (4,0) (5,0) (6,0) (7,0) (8,0) (9,0) (10,0) (11,0) (12,0) (13,0) (14,0) (15,0) (16,0) (17,0) (18,0) (19,0) (20,0)
    }; 
    \addplot[draw=tabgreen,fill=tabgreen] coordinates {
    (1,0) (2,0) (3,0) (4,0) (5,0) (6,0) (7,0) (8,0) (9,0) (10,0) (11,0) (12,0) (13,0) (14,0) (15,0) (16,0) (17,0) (18,0) (19,0) (20,0)
    }; 
    \addplot[draw=tabred,fill=tabred] coordinates {
    (1,497) (2,500) (3,500) (4,500) (5,500) (6,500) (7,500) (8,500) (9,500) (10,500) (11,500) (12,500) (13,500) (14,500) (15,500) (16,500) (17,500) (18,500) (19,500) (20,500)
    }; 
    \addplot[draw=tabpurple,fill=tabpurple] coordinates {
    (1,3) (2,0) (3,0) (4,0) (5,0) (6,0) (7,0) (8,0) (9,0) (10,0) (11,0) (12,0) (13,0) (14,0) (15,0) (16,0) (17,0) (18,0) (19,0) (20,0)
    }; 
  \end{groupplot}
\end{tikzpicture}
        \caption{\textit{Routing visualization.} We plot the decisions made by the router $\mathcal{R}$ over the depth of the network.}
        \label{fig:exp:routing}
    \end{minipage}%
    \hfill
    \begin{minipage}[b]{0.49\textwidth}
        \begin{tikzpicture}
  \begin{groupplot}[
  group style={group size=4 by 1, horizontal sep=0.35cm,},
  width=0.4\textwidth, height=0.575\textwidth,
  xtick pos=left, ytick pos=left, xlabel near ticks, ylabel near ticks, xtick=data, ymajorgrids=true,
  grid, grid style=dashed, tick label style={font=\tiny}, title style={font=\tiny, anchor=center, at={(0.5,1)}}, label style={font=\tiny}, tick align=inside,
  axis on top, enlargelimits=0.1,     
  xtick={0, 1, 2, 3}, xticklabels={.25, .5, .9, .99}, yticklabels=\empty,
  ]

    \nextgroupplot[
    ylabel={PSNR (dB)}, ylabel style={anchor=center, at={(-0.1,0.5)},},
    xlabel={Weight $\alpha$}, every axis x label/.append style={at=(ticklabel cs:2.25)},
    title={\textit{Dehazing}},
    ]
    \addplot[color=tabblue, mark=*]
    coordinates {
    (0,31.48)(1,32.06)(2,32.30)(3,31.95)
    };

    \nextgroupplot[title={\textit{Deraining}},]
    \addplot[color=taborange, mark=*]
    coordinates {
    (0,37.17)(1,36.99)(2,37.10)(3,36.79)
    };

    \nextgroupplot[title={\textit{Denoising}},]
    \addplot[color=tabgreen, mark=*]
    coordinates {
    (0,33.92)(1,33.91)(2,33.92)(3,33.89)
    };

    \nextgroupplot[title={\textit{Average}},]
    \addplot[color=tabred, mark=*]
    coordinates {
    (0,32.36)(1,32.44)(2,32.51)(3,32.37)
    };

  \end{groupplot}
\end{tikzpicture}            
        \caption{\textit{Impact of $\alpha$.} We demonstrate the impact of adjusting the influence of encoder parameters on the controller by varying the parameter $\alpha$.}
        \label{fig:exp:ema_alpha}
    \end{minipage}
\end{figure}

\begin{figure}[htbp]
        \begin{tblr}{
          colspec = {@{}lX[c]X[c]X[c]X[c]X[c]X[c]@{}},
            colsep=0.1pt,
          rows={rowsep=0.75pt},
          stretch = 0,
        }
        \rotatebox{90}{\textcolor{white}{...}Input} &  
        \includegraphics[width=0.156\textwidth, height=0.1\textwidth]{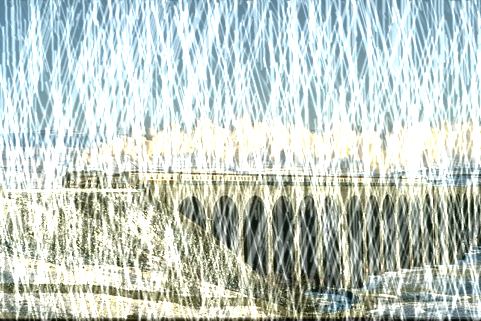} & \includegraphics[width=0.156\textwidth, height=0.1\textwidth,]{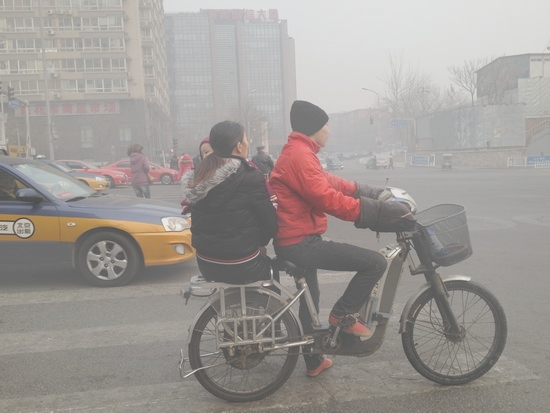} & 
        \includegraphics[width=0.156\textwidth, height=0.1\textwidth]{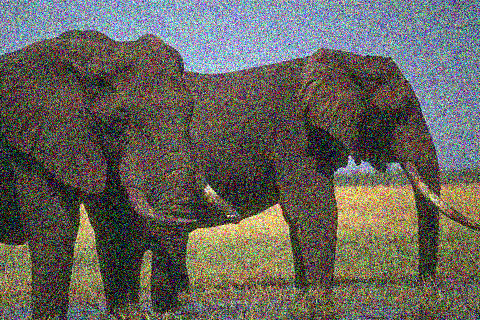}
        & \includegraphics[width=0.156\textwidth, height=0.1\textwidth]{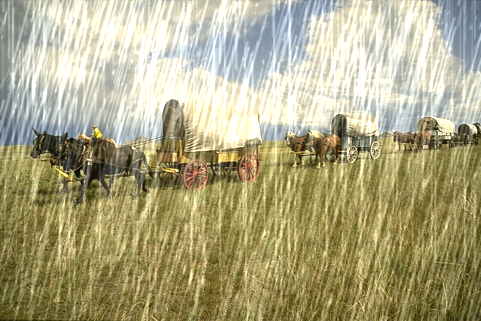} & 
        \includegraphics[width=0.156\textwidth, height=0.1\textwidth]{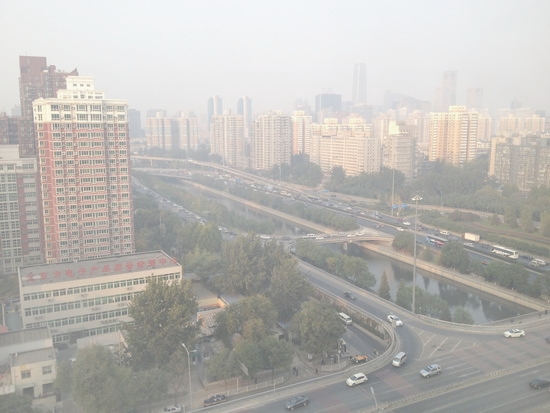} & 
        \includegraphics[width=0.156\textwidth, height=0.1\textwidth]{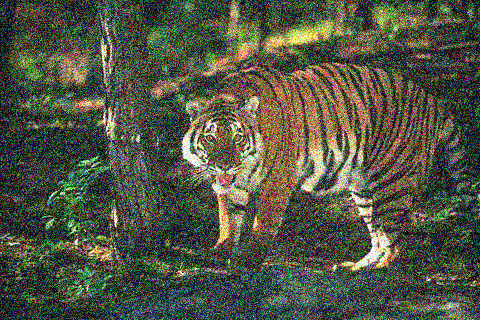}\\
        
        \rotatebox{90}{\textcolor{white}{.....}$\mathcal{E}_{A}$} & 
        \includegraphics[width=0.156\textwidth, height=0.1\textwidth]{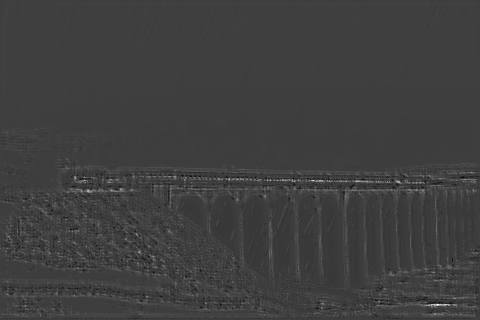} & \includegraphics[width=0.156\textwidth, height=0.1\textwidth]{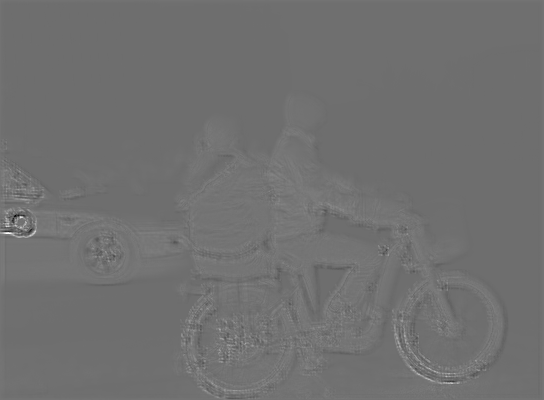} & 
        \includegraphics[width=0.156\textwidth, height=0.1\textwidth]{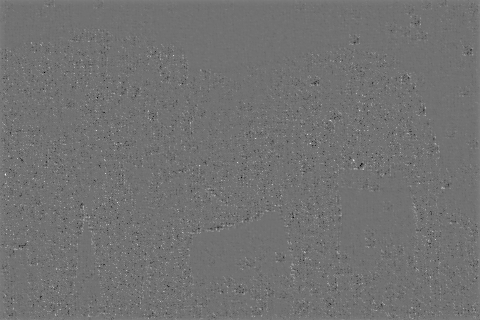} & \includegraphics[width=0.156\textwidth, height=0.1\textwidth]{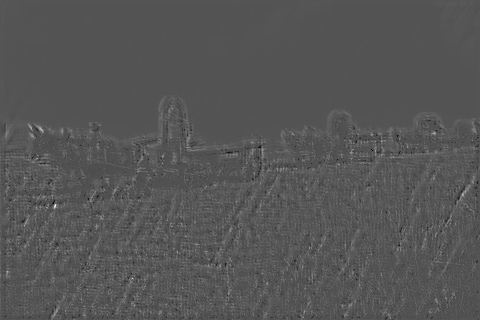} & 
        \includegraphics[width=0.156\textwidth, height=0.1\textwidth]{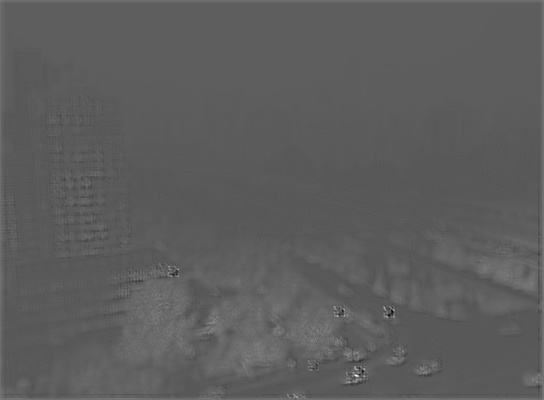} & 
        \includegraphics[width=0.156\textwidth, height=0.1\textwidth]{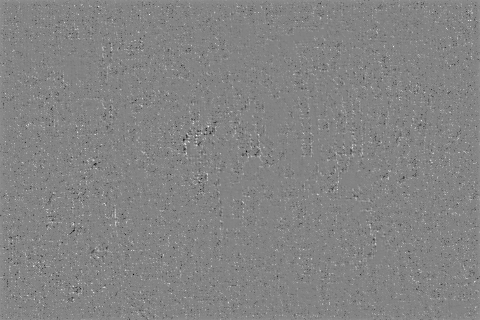}\\

        \rotatebox{90}{\textcolor{white}{.....}$\mathcal{E}_{D}$} & 
        \includegraphics[width=0.156\textwidth, height=0.1\textwidth]{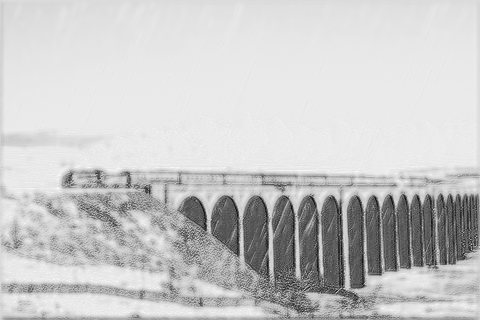} & \includegraphics[width=0.156\textwidth, height=0.1\textwidth]{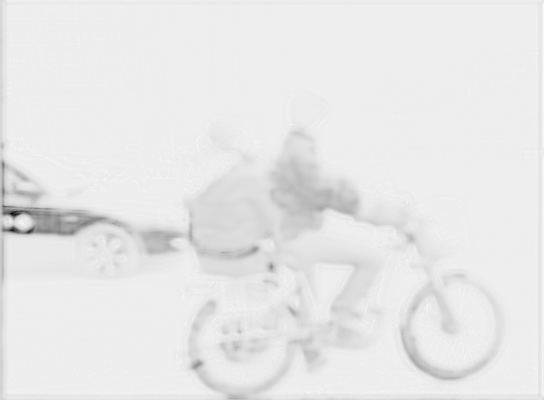} & 
        \includegraphics[width=0.156\textwidth, height=0.1\textwidth]{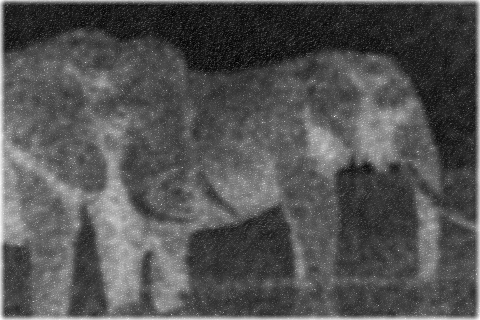} & \includegraphics[width=0.156\textwidth, height=0.1\textwidth]{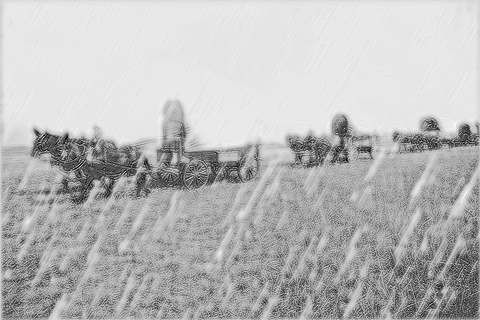} & 
        \includegraphics[width=0.156\textwidth, height=0.1\textwidth]{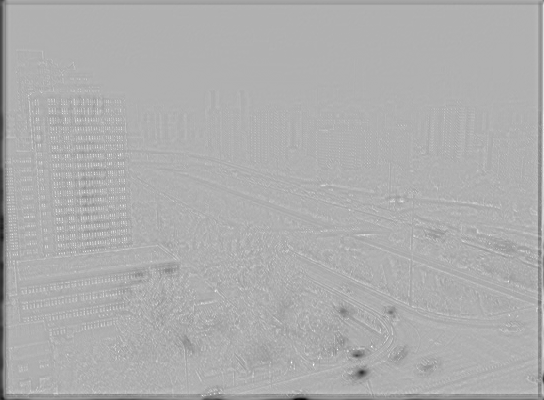} & 
        \includegraphics[width=0.156\textwidth, height=0.1\textwidth]{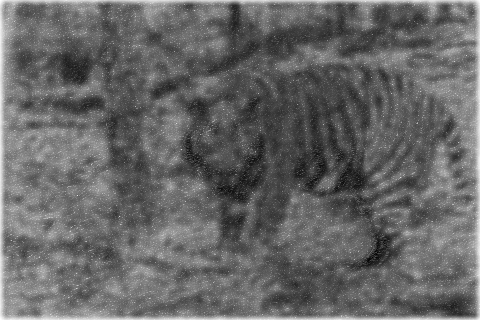}\\

        \rotatebox{90}{\textcolor{white}{.....}$\mathcal{E}_{C}$} & 
        \includegraphics[width=0.156\textwidth, height=0.1\textwidth]{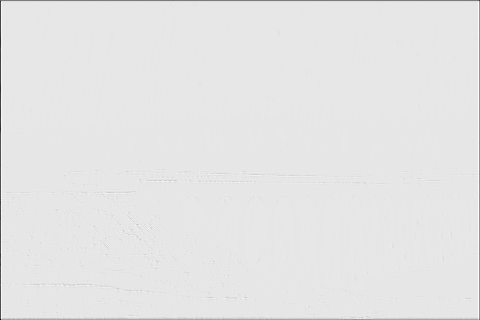} & \includegraphics[width=0.156\textwidth, height=0.1\textwidth]{figures/feat_vis/deraining/implicit/controllerrain-004_39.68.png} & 
        \includegraphics[width=0.156\textwidth, height=0.1\textwidth]{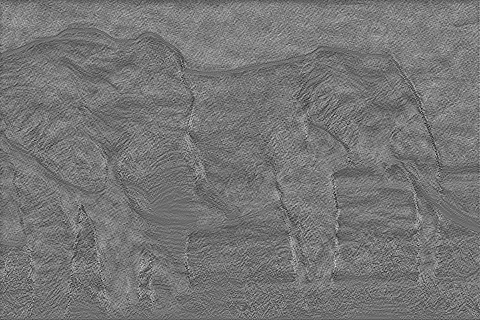} & \includegraphics[width=0.156\textwidth, height=0.1\textwidth]{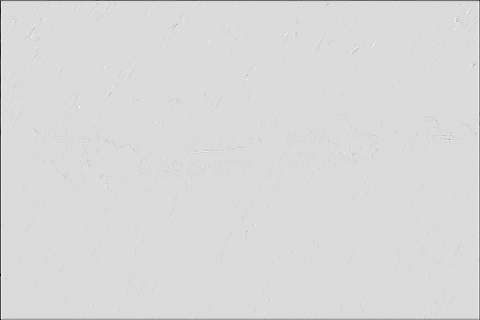} & 
        \includegraphics[width=0.156\textwidth, height=0.1\textwidth]{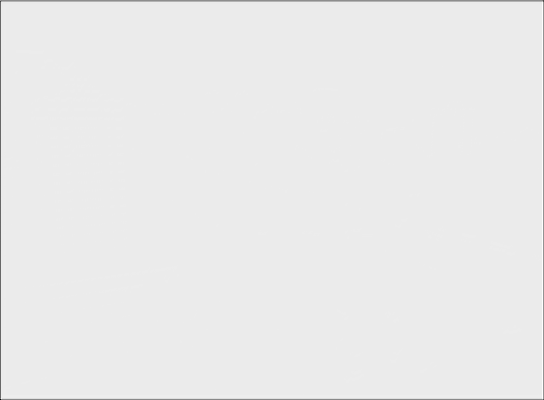} & 
        \includegraphics[width=0.156\textwidth, height=0.1\textwidth]{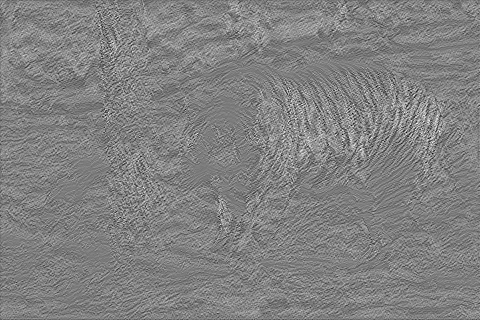}\\
        \end{tblr}

    \caption{\textit{Feature visualization.} We visualize the features learned by the agnostic expert $\mathcal{E}_{A}$ and degradation-specific experts $\mathcal{E}_{D}$. The controller $\mathcal{E}_{C}$ adeptly captures degradation-specific patterns, revealing distinct features for various types of corruptions. Zoom in for better view.}
  \label{fig:exp:feature_vis}
\end{figure}

\paragraph{Visual Results.}
We show visual results obtained under the three degradation settings in \cref{fig:exp:visual_results}. 
In certain demanding hazy scenarios, as demonstrated, in \cref{fig:exp:visual_results}, both AirNet~\cite{li2022airnet} and PromptIR~\cite{potlapalli2023promptir} reveal constraints in completely eradicating haziness, resulting in noticeable color intensity discrepancies, whereas our approach ensures precise color reconstruction. Furthermore, in challenging rainy scenes, these popular approaches continue to exhibit notable remnants of rain, which our method adeptly eliminates, distinguishing itself from other approaches. Additionally, our method also generates clear and sharp denoised outputs. Coupled with the quantitative comparisons, these findings underscore the effectiveness of our method.

\subsection{Ablation Studies}
We conduct detailed studies on the components within our method. All experiments are conducted in the All-in-One setting with three degradations.

\paragraph{Architecture Contribution.}
As detailed in \cref{tab:exp:high_level_ablations}, we assess the efficacy of our proposed key architectural components by contrasting them with a baseline method devoid of our modules. This baseline adopts a scaled-down version of the architecture found in Restormer~\cite{Zamir2021Restormer}. Successively integrating the proposed experts and controller into the baseline architecture yields a significant and consistent enhancement. Overall, our framework achieves a notable average improvement of $0.65$ dB, attributable to the efficacy of our proposed components and we make the following observations: (i) It is difficult for the model to learn about the degradations without the assistance of the experts, (ii) Incorporation of the controller seems to be beneficial for the overall network learning (w/o controller, the framework's reconstruction fidelity decreases by an average of $0.28$ dB), and (iii) The routing strategy is crucial for the overall performance improvement. The quantitative results in \cref{tab:exp:high_level_ablations} \textit{(d)} and the routing visualization in \cref{fig:exp:routing} highlight the critical role of explicitly correlating routing with the target degradation.

\paragraph{Self-learnable Control.}
The controller module leverages the accumulated prior knowledge to generate degradation-aware features. It primarily focuses on the most affected regions by updating its parameters via an exponential moving average from the encoder-side agnostic expert which develops an intrinsic understanding of degradations during optimization. We ablate the selection of $\alpha$, with results visualized in \cref{fig:exp:ema_alpha}. Our findings reveal that reconstruction performance is sensitive to $\alpha$, with $\alpha=0.9$ empirically yielding the best average results and as already discussed above with the controller, the overall network yields higher performance gain. This demonstrates that leveraging "self-knowledge" significantly enhances the performance of the restorer without incurring additional costs, as agnostic representations are already learned during training. We further visualize and discuss the intricacies learned by the controller in \cref{fig:exp:feature_vis} and \cref{sec:discussion}, respectively.

\paragraph{Model efficiency.}
\cref{tab:exp:efficiency_overview} compares memory usage, GMACS, and model parameters, demonstrating our framework's superiority over state-of-the-art All-in-One restorers. By employing an scaled-down Transformer architecture and parameter-efficient experts that process features in a low-rank space, we achieve reductions of $84\%$ in GMACS and $66\%$ in memory consumption compared to PromptIR~\cite{potlapalli2023promptir} on an image size $224 \times 224$.

\section{Discussion}
\label{sec:discussion}

Our framework incorporates specialized experts at various levels, each dedicated to distinct facets of the restoration problem. These experts either implicitly model degradation specifics, process task-agnostic information or leverage accumulated prior knowledge to control the restoration process. This section aims to elucidate the specific expertise of each component.

\paragraph{Expertized degradation learning.}
The decision-making process of our degradation-aware routers is illustrated in \cref{fig:exp:routing} (for five degradations). In the absence of the auxiliary degradation classification loss, the selection of experts at each layer does not correspond effectively to the actual type of input degradation. This misalignment results in a more randomized selection, ultimately leading to diminished restoration quality. Although the specific assignment of experts to degradations is arbitrary, maximizing their effectiveness requires a robust router, facilitating the learning of degradation-dependent representations, thereby harnessing the full potential of the parameter-efficient experts. As illustrated in the \cref{fig:exp:feature_vis}, the agnostic expert capture general patterns, such as high-frequency details, across different degradations, whereas the specialized experts concentrates strongly on task-related information.

\paragraph{Controller captures degradation characteristics.}
To underscore the significance of the proposed controller module, we analyze its learned representations, as illustrated in \cref{fig:exp:feature_vis}. This analysis vividly demonstrates the controller's adeptness in capturing degradation patterns, elucidating regions afflicted by severe corruption. 
For instance, in the case of dehazing, where the degradation predominantly impacts the lower frequency spectrum, the controller features manifest a homogeneous distribution. Conversely, while noise presents as stochastic fluctuations in pixel values, impacting higher frequencies, the learned controller features effectively grasps these stochastic variations.

\section{Conclusion}
This paper introduces DaAIR, an efficient image restoration model that can deal with any type of degradation. Our model sets a new standard in degradation awareness, achieving unparalleled efficiency and fidelity in All-in-One image restoration. 
In our approach, we intricately design an innovative degradation-aware learner to extract inherent commonalities and capture subtle nuances across various degradations, collaboratively developing a comprehensive degradation-aware representation in the low-rank regime. 
Specifically, by dynamically allocating the explicit experts to the input degradation, we further enhance the capability of our model. Additionally, we further demonstrate that the degradation-aware representation serves as an effective auxiliary supervisory signal within our method, significantly enhancing the restoration fidelity. Extensive experiments on All-in-One image restoration across diverse degradations reveal that our proposed DaAIR consistently outperforms recent state-of-the-art methods, pushing the boundaries of efficient All-in-One restoration and affirming its practicality.

\bibliography{main}
\bibliographystyle{abbrvnat}

\newpage
\appendix
\begin{figure}[t]
    \centering
    \footnotesize
    \begin{subfigure}{\textwidth}
    \begin{tblr}{
      colspec = {@{}lX[c]X[c]X[c]X[c]X[c]@{}},
      colsep=0.1pt,
      rows={rowsep=0.75pt},
      stretch = 0,
    }
     &  Input & AirNet~\cite{li2022airnet} & PromptIR~\cite{potlapalli2023promptir} & Ours & GT \\
        \SetCell[r=3]{l}{\rotatebox{90}{Dehazing}} &       
        \includegraphics[width=0.19\textwidth, height=0.12\textwidth]{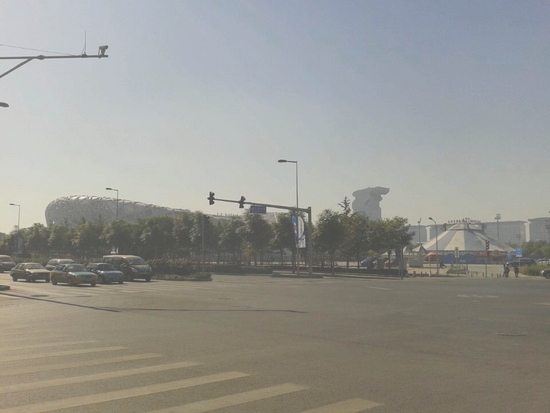} & \includegraphics[width=0.19\textwidth, height=0.12\textwidth]{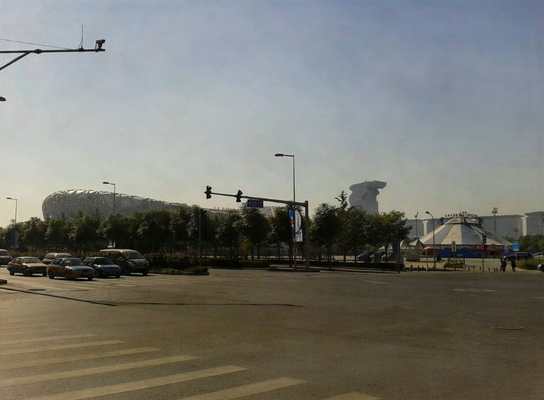} & 
        \includegraphics[width=0.19\textwidth, height=0.12\textwidth]{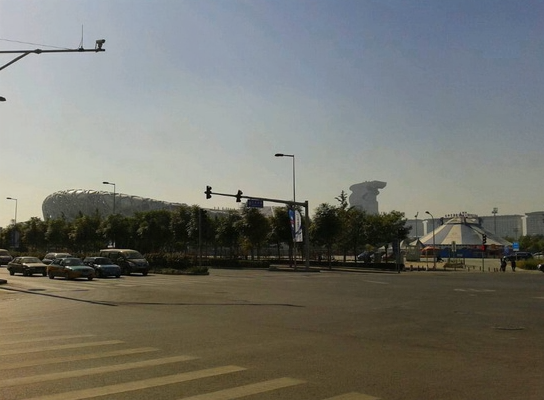} & \includegraphics[width=0.19\textwidth, height=0.12\textwidth]{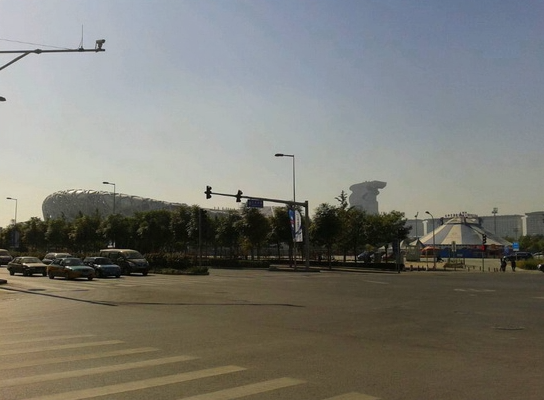} & 
        \includegraphics[width=0.19\textwidth, height=0.12\textwidth]{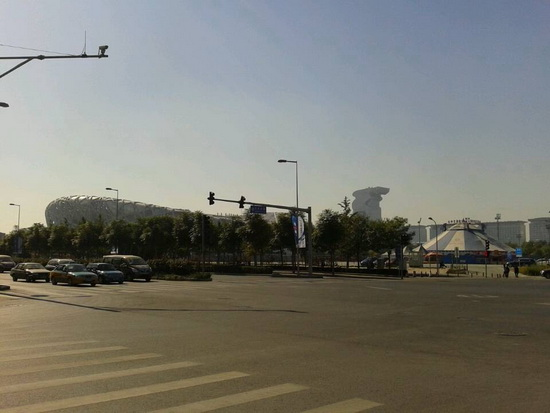}\\

        & \includegraphics[width=0.19\textwidth, height=0.12\textwidth]{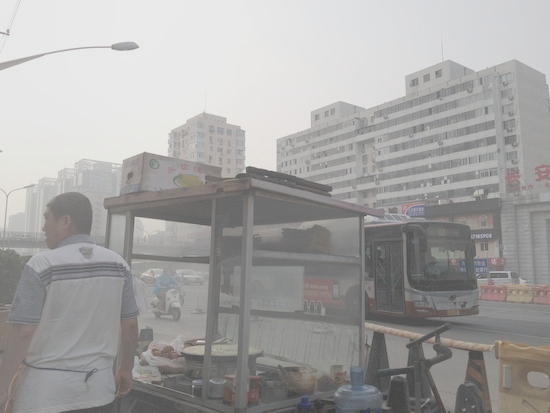} & \includegraphics[width=0.19\textwidth, height=0.12\textwidth]{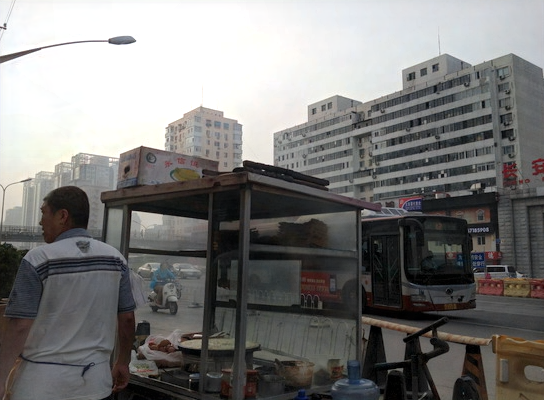} & 
        \includegraphics[width=0.19\textwidth, height=0.12\textwidth]{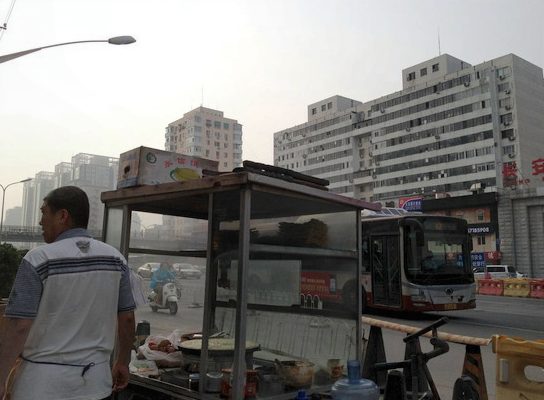} & \includegraphics[width=0.19\textwidth, height=0.12\textwidth]{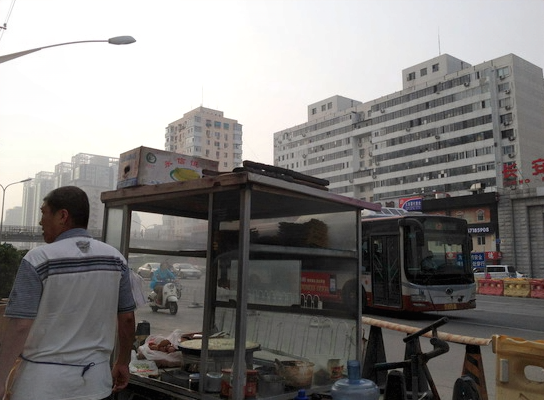} & 
        \includegraphics[width=0.19\textwidth, height=0.12\textwidth]{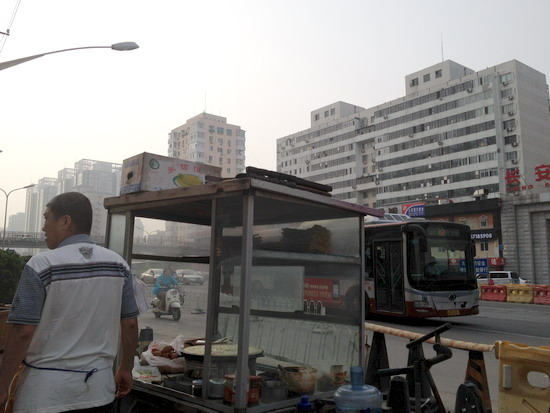} \\

        & \includegraphics[width=0.19\textwidth, height=0.12\textwidth]{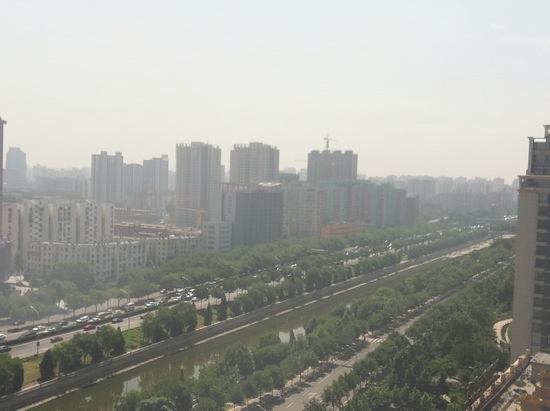} & \includegraphics[width=0.19\textwidth, height=0.12\textwidth, ]{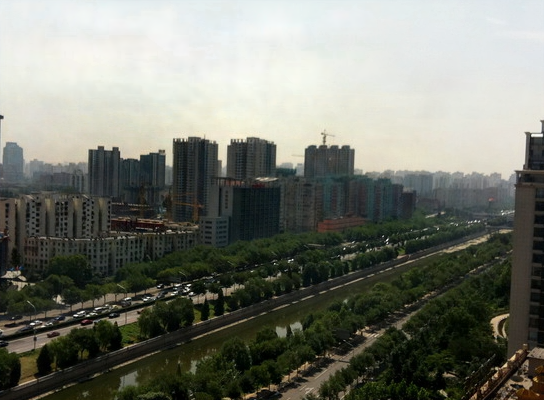} & 
        \includegraphics[width=0.19\textwidth, height=0.12\textwidth,]{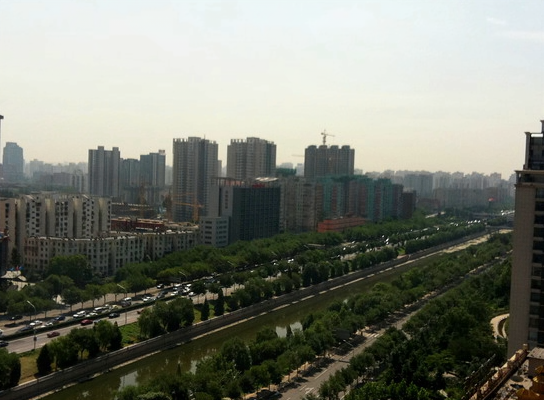} & 
         \includegraphics[width=0.19\textwidth, height=0.12\textwidth,]{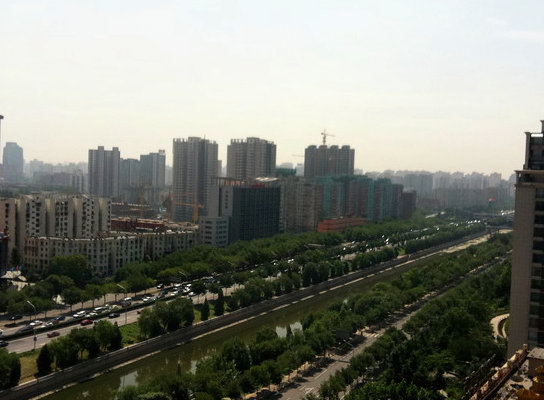} & 
         \includegraphics[width=0.19\textwidth, height=0.12\textwidth,]{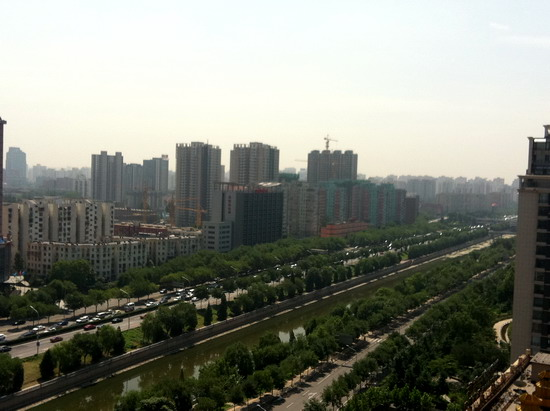}\\

    \end{tblr}
    \end{subfigure}
\hfill
    \begin{subfigure}{\textwidth}
        \begin{tblr}{
          colspec = {@{}lX[c]X[c]X[c]X[c]X[c]@{}},
          colsep=0.1pt,
          rows={rowsep=0.75pt},
          stretch = 0,
    }
       \SetCell[r=3]{l}{\rotatebox{90}{Deraining}} & 
       \includegraphics[width=0.19\textwidth, height=0.12\textwidth]{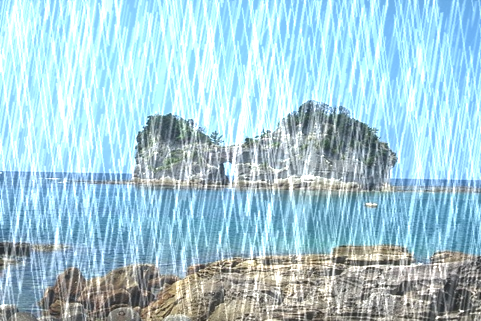} & \includegraphics[width=0.19\textwidth, height=0.12\textwidth,]{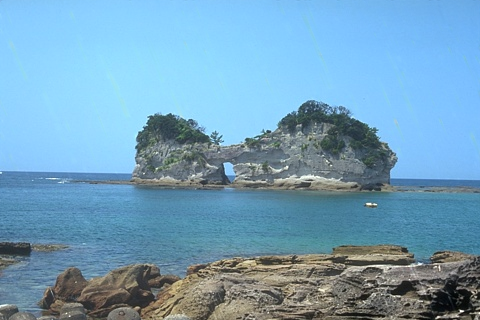} & 
       \includegraphics[width=0.19\textwidth, height=0.12\textwidth,]{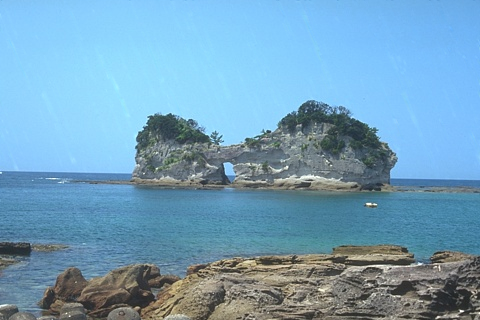} & \includegraphics[width=0.19\textwidth, height=0.12\textwidth,]{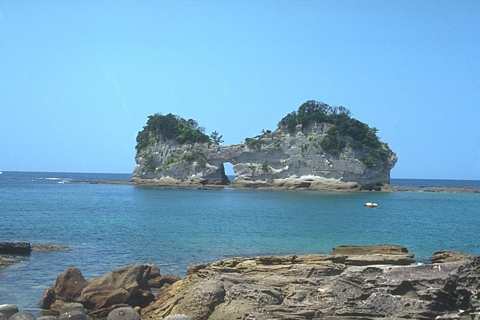} & 
       \includegraphics[width=0.19\textwidth, height=0.12\textwidth, ]{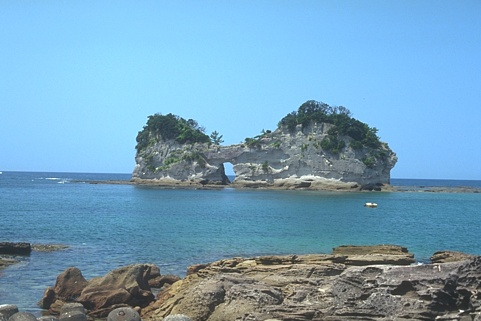}\\

       & \includegraphics[width=0.19\textwidth, height=0.12\textwidth]{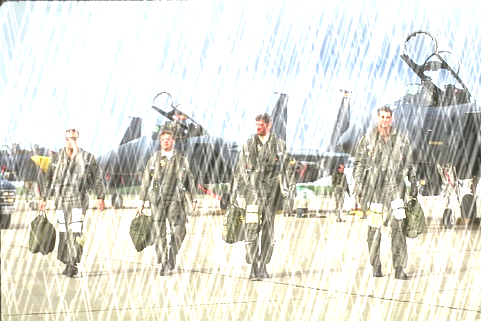} & \includegraphics[width=0.19\textwidth, height=0.12\textwidth,]{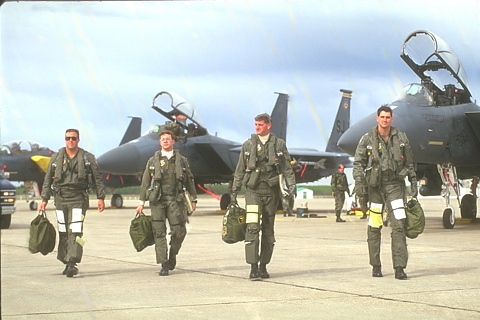} & 
       \includegraphics[width=0.19\textwidth, height=0.12\textwidth,]{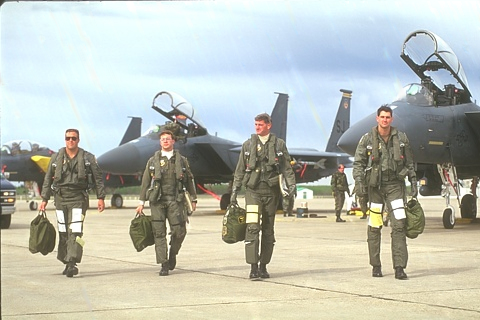} & \includegraphics[width=0.19\textwidth, height=0.12\textwidth,]{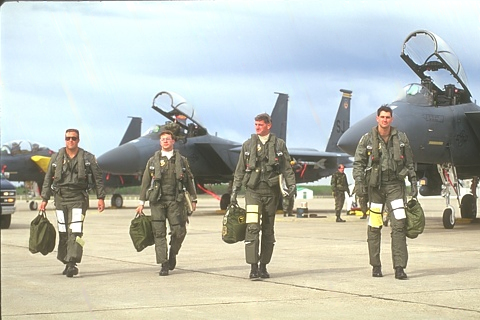} & 
       \includegraphics[width=0.19\textwidth, height=0.12\textwidth, ]{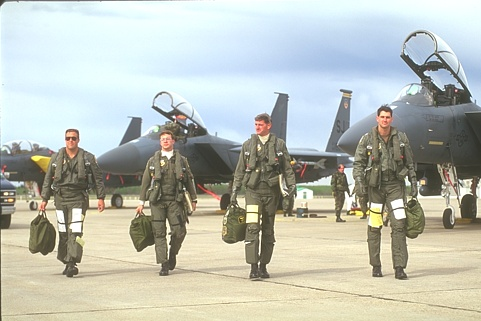}\\

       & \includegraphics[width=0.19\textwidth, height=0.12\textwidth]{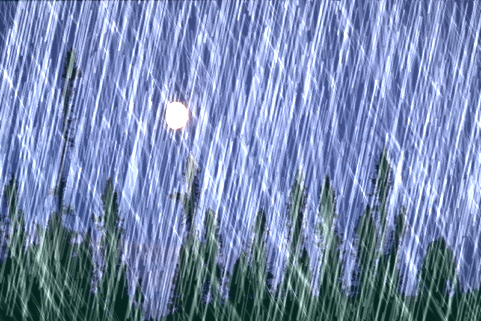} & \includegraphics[width=0.19\textwidth, height=0.12\textwidth]{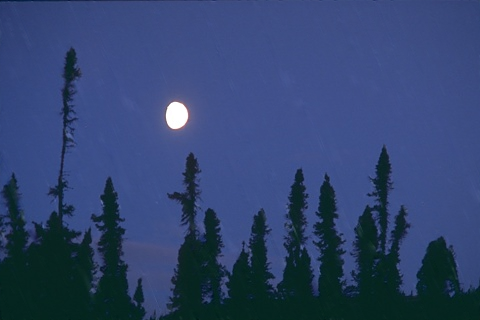} & 
       \includegraphics[width=0.19\textwidth, height=0.12\textwidth]{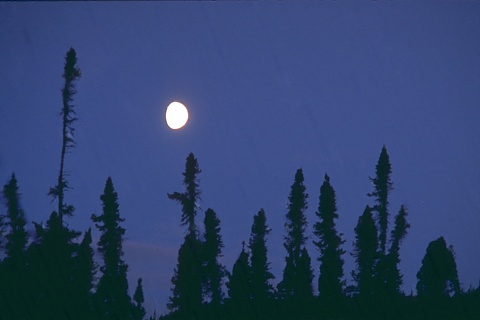} & \includegraphics[width=0.19\textwidth, height=0.12\textwidth]{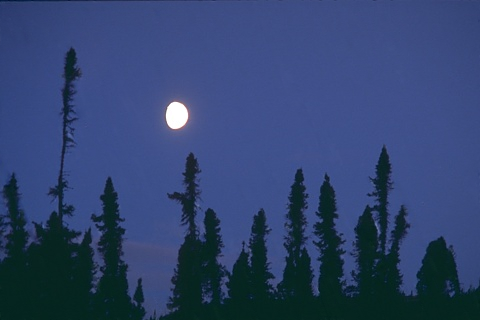} & 
       \includegraphics[width=0.19\textwidth, height=0.12\textwidth]{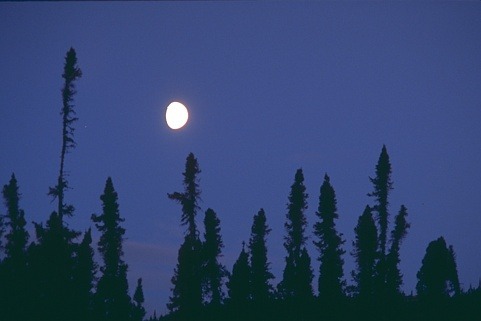}\\
    \end{tblr}
    \end{subfigure}
\hfill
    \begin{subfigure}{\textwidth}
        \begin{tblr}{
          colspec = {@{}lX[c]X[c]X[c]X[c]X[c]@{}},
          colsep=0.1pt,
          rows={rowsep=0.75pt},
          stretch = 0,
        }
        \SetCell[r=3]{l}{\rotatebox{90}{Denoising}} &  
        \includegraphics[width=0.19\textwidth, height=0.12\textwidth]{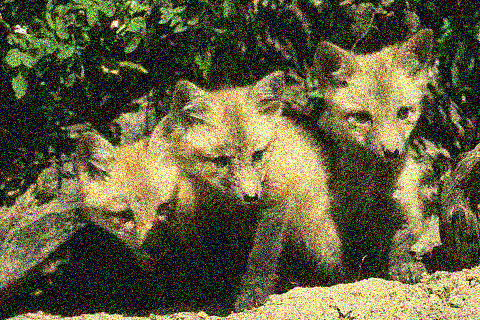} & \includegraphics[width=0.19\textwidth, height=0.12\textwidth,]{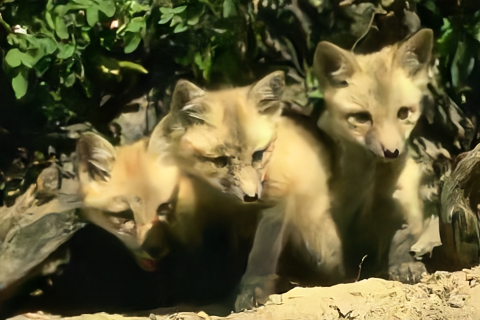} & 
        \includegraphics[width=0.19\textwidth, height=0.12\textwidth]{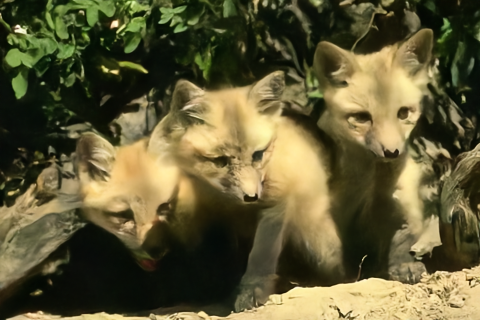}
        & \includegraphics[width=0.19\textwidth, height=0.12\textwidth]{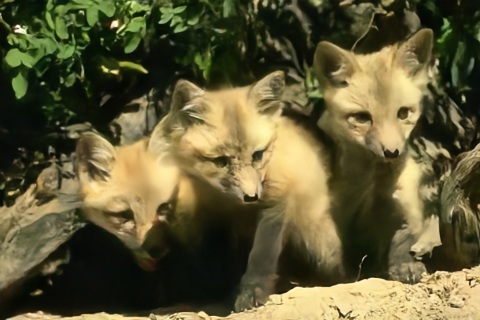} & 
        \includegraphics[width=0.19\textwidth, height=0.12\textwidth]{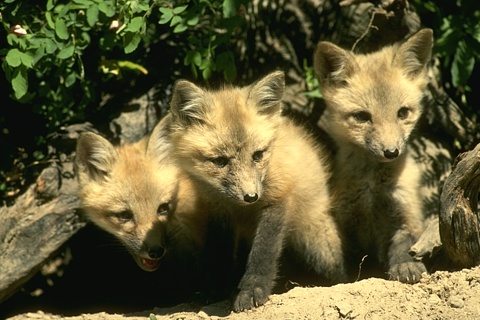}\\

        &  \includegraphics[width=0.19\textwidth, height=0.12\textwidth]{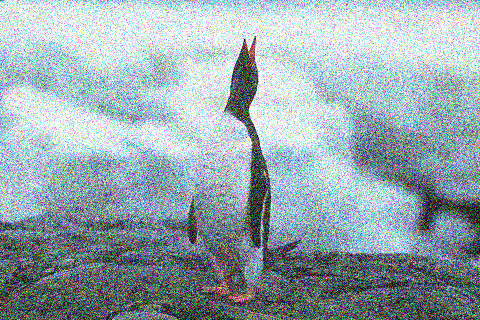} & \includegraphics[width=0.19\textwidth, height=0.12\textwidth]{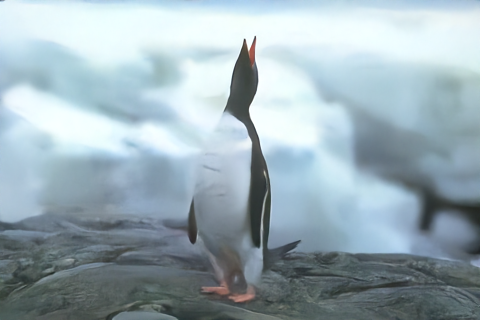} & 
        \includegraphics[width=0.19\textwidth, height=0.12\textwidth]{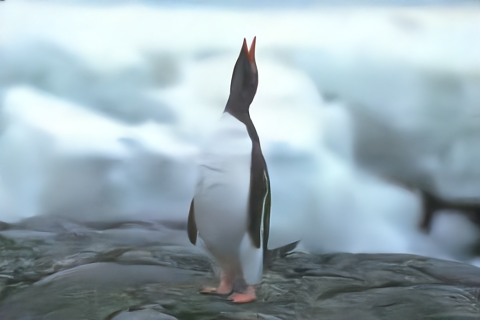} 
        & \includegraphics[width=0.19\textwidth, height=0.12\textwidth]{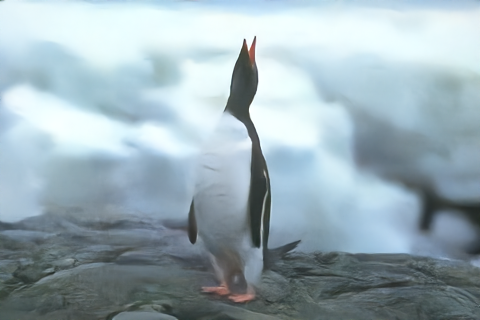} & 
        \includegraphics[width=0.19\textwidth, height=0.12\textwidth]{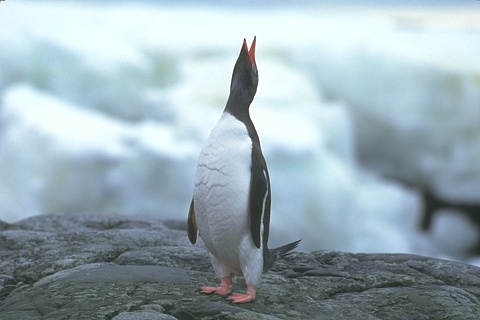}\\
        
        &  \includegraphics[width=0.19\textwidth, height=0.12\textwidth]{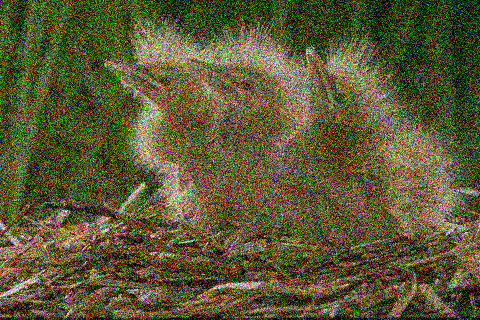} & \includegraphics[width=0.19\textwidth, height=0.12\textwidth]{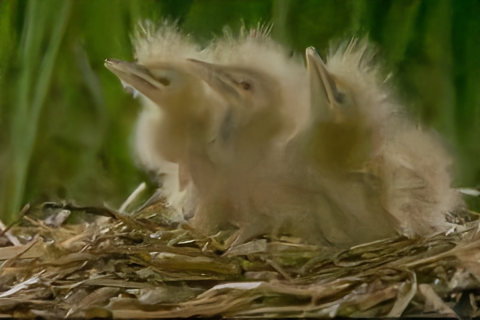} & 
        \includegraphics[width=0.19\textwidth, height=0.12\textwidth]{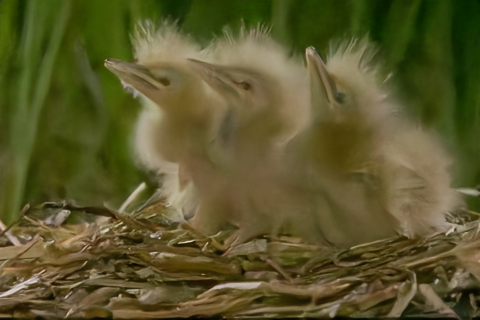} & \includegraphics[width=0.19\textwidth, height=0.12\textwidth]{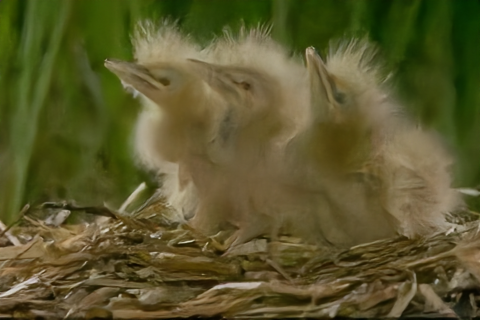} & 
        \includegraphics[width=0.19\textwidth, height=0.12\textwidth]{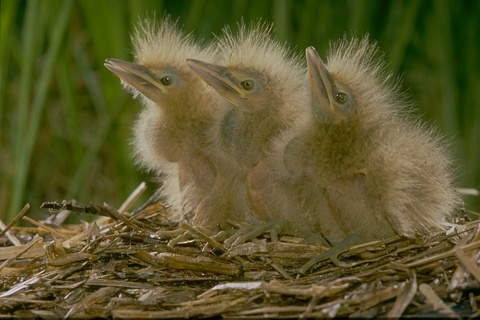}\\

        \end{tblr}

    \end{subfigure}
    \caption{Visual comparison of DaAIR with state-of-the-art methods on challenging cases for the All-in-One setting considering three degradations.}
    \label{fig:supp:visual_results}
\end{figure}

\begin{figure}[t]
        \begin{tblr}{
          colspec = {@{}lX[c]X[c]X[c]X[c]X[c]X[c]@{}},
            colsep=0.1pt,
          rows={rowsep=0.75pt},
          stretch = 0,
        }
        \rotatebox{90}{\textcolor{white}{...}Input} &  
        \includegraphics[width=0.156\textwidth, height=0.1\textwidth]{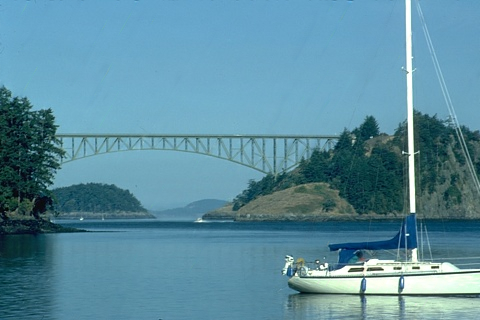} 
        & \includegraphics[width=0.156\textwidth, height=0.1\textwidth,]{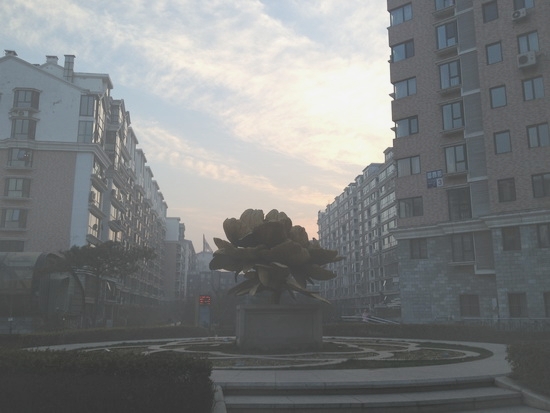} & 
        \includegraphics[width=0.156\textwidth, height=0.1\textwidth]{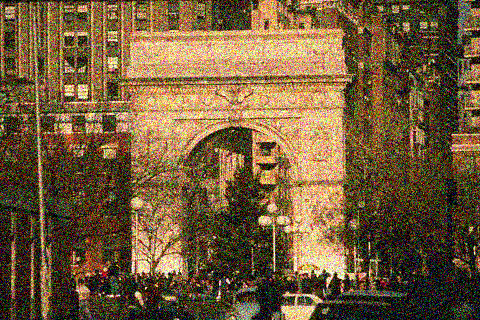}
        & \includegraphics[width=0.156\textwidth, height=0.1\textwidth]{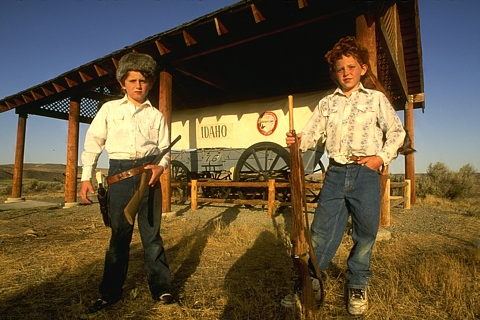} & 
        \includegraphics[width=0.156\textwidth, height=0.1\textwidth]{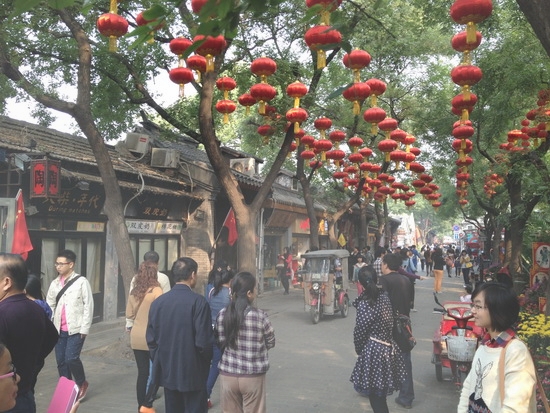} & 
        \includegraphics[width=0.156\textwidth, height=0.1\textwidth]{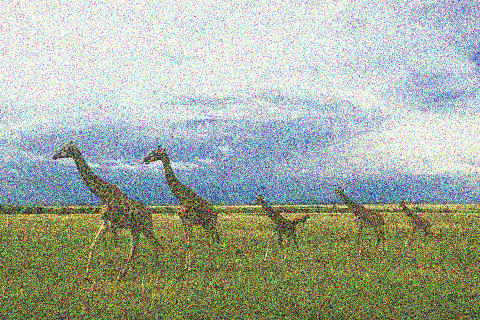}\\
        
        \rotatebox{90}{\textcolor{white}{.....}$\mathcal{E}_{A}$} & 
        \includegraphics[width=0.156\textwidth, height=0.1\textwidth]{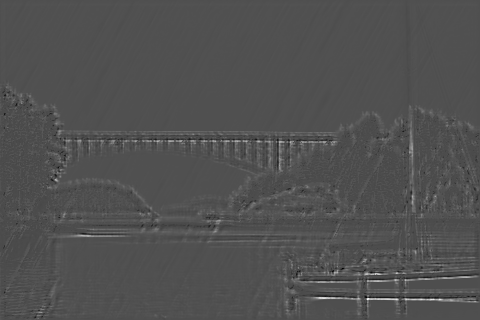}
        & \includegraphics[width=0.156\textwidth, height=0.1\textwidth]{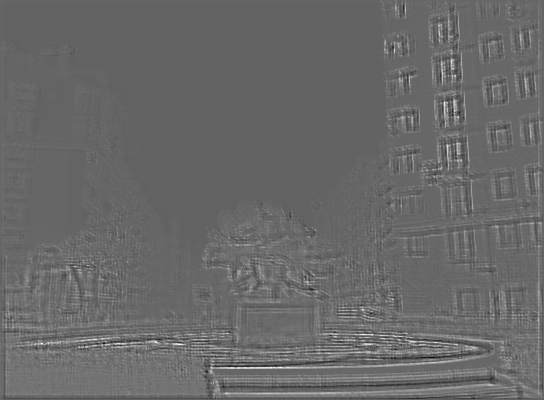} & 
        \includegraphics[width=0.156\textwidth, height=0.1\textwidth]{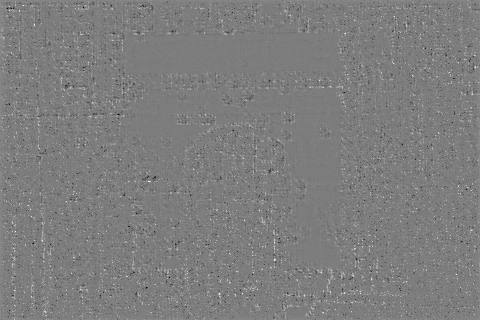} & \includegraphics[width=0.156\textwidth, height=0.1\textwidth]{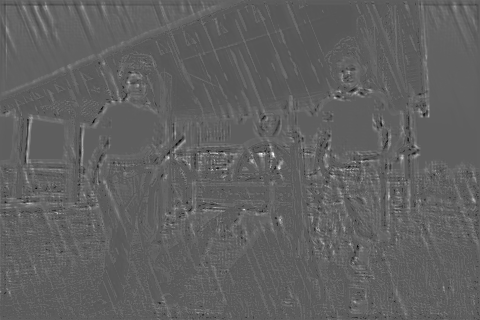} & 
        \includegraphics[width=0.156\textwidth, height=0.1\textwidth]{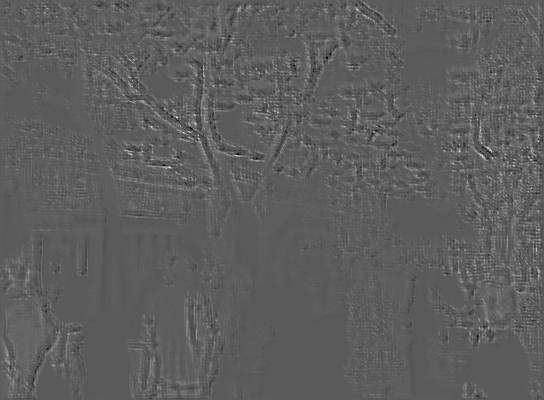} & 
        \includegraphics[width=0.156\textwidth, height=0.1\textwidth]{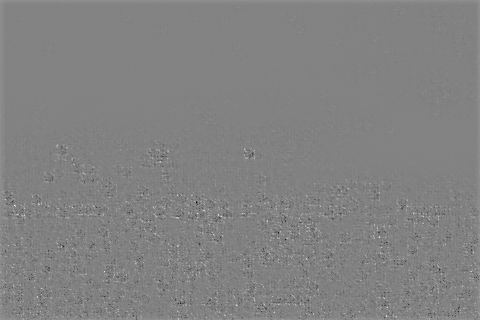}\\

        \rotatebox{90}{\textcolor{white}{.....}$\mathcal{E}_{D}$} & 
        \includegraphics[width=0.156\textwidth, height=0.1\textwidth]{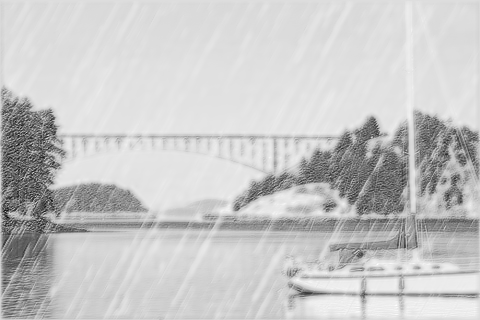} &
        \includegraphics[width=0.156\textwidth, height=0.1\textwidth]{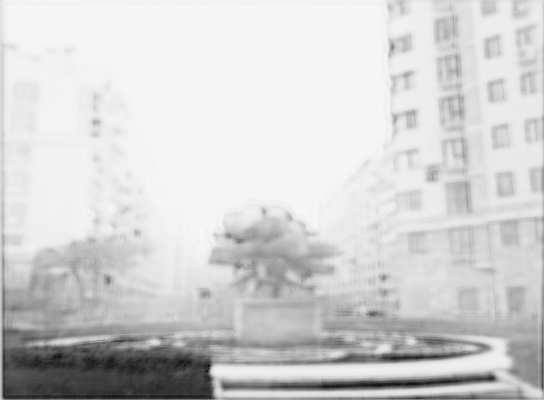} & 
        \includegraphics[width=0.156\textwidth, height=0.1\textwidth]{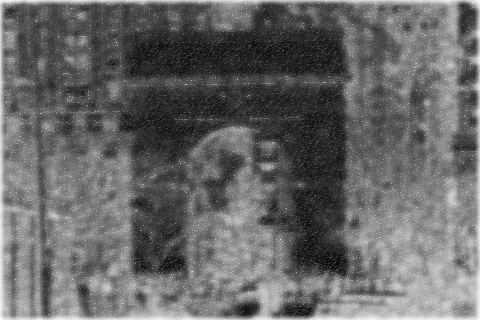} & \includegraphics[width=0.156\textwidth, height=0.1\textwidth]{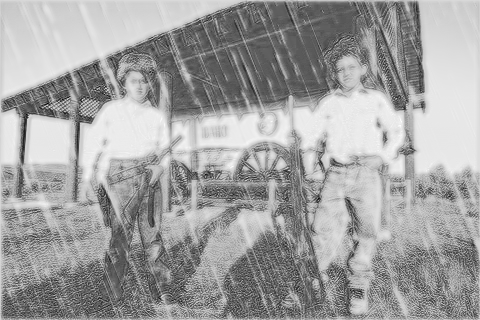} & 
        \includegraphics[width=0.156\textwidth, height=0.1\textwidth]{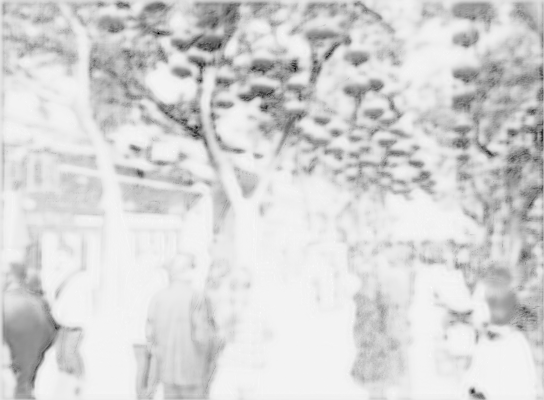} & 
        \includegraphics[width=0.156\textwidth, height=0.1\textwidth]{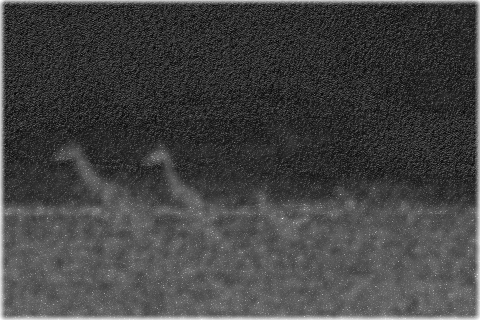}\\

        \rotatebox{90}{\textcolor{white}{.....}$\mathcal{E}_{C}$} & 
        \includegraphics[width=0.156\textwidth, height=0.1\textwidth]{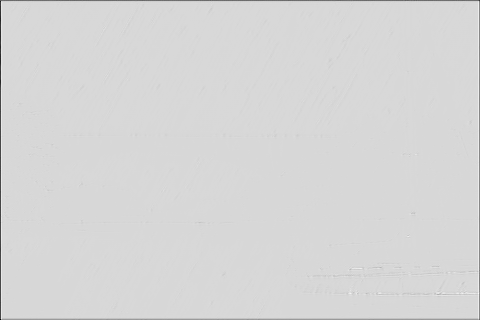} & 
        \includegraphics[width=0.156\textwidth, height=0.1\textwidth]{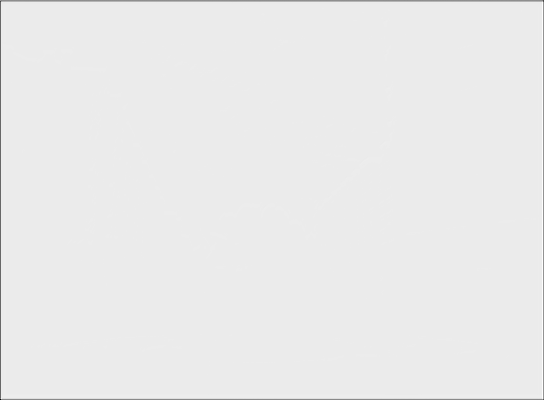} & 
        \includegraphics[width=0.156\textwidth, height=0.1\textwidth]{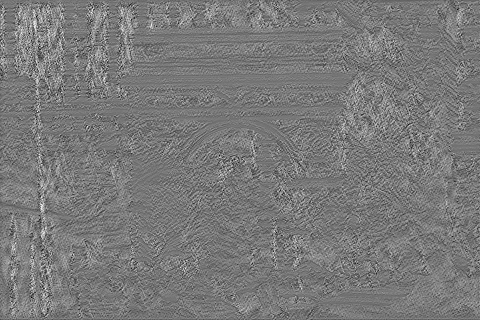} & 
        \includegraphics[width=0.156\textwidth, height=0.1\textwidth]{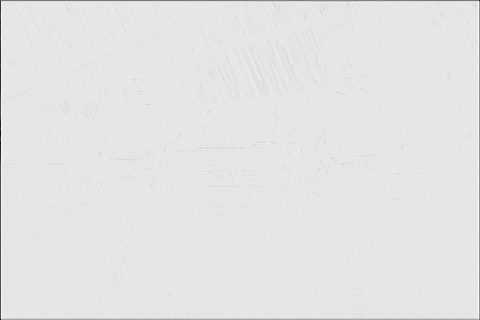} & 
        \includegraphics[width=0.156\textwidth, height=0.1\textwidth]{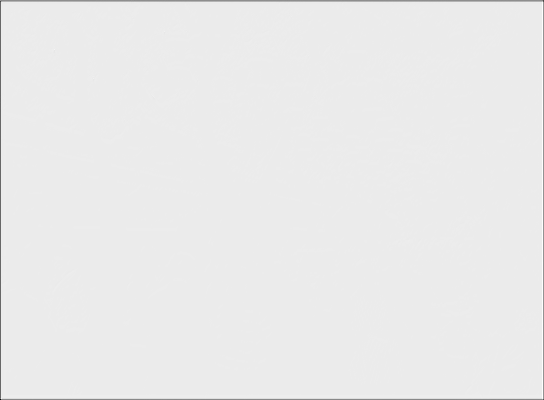} & 
        \includegraphics[width=0.156\textwidth, height=0.1\textwidth]{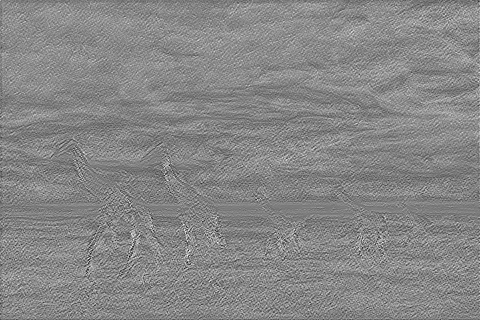}\\
        \end{tblr}

    \caption{\textit{More feature visualization.} We visualize the features learned by the agnostic expert $\mathcal{E}_{A}$, degradation-specific experts $\mathcal{E}_{D}$ and the controller $\mathcal{E}_{C}$. Zoom in for better view.}
    \label{fig:supp:feat_visuals}
\end{figure}

\section{Further Implementation Details}
Across all experiments, we maintain a consistent random seed to ensure reproducibility. Our implementation was built upon the publicly available PyTorch-based \textit{PromptIR} codebase, leveraging it for architecture development and training. Additionally, we employed the \textit{fvcore} Python package to compute GMACS and parameter counts.
\paragraph{Baseline for architecture ablation.}
In this section, we elaborate on the baseline method utilized for the ablation study presented in \cref{tab:exp:high_level_ablations}. Our model is constructed upon the Restormer architecture~\cite{Zamir2021Restormer}, akin to previous All-in-One models~\cite{potlapalli2023promptir,dudhane2024dynet}. However, we implement modifications by reducing the number of blocks within each level of the encoder and decoder, while initializing the input embedding size to $32$, progressively doubling it in subsequent levels. Furthermore, within our final model, we replace the channel-wise self-attention layer on the decoder side of the UNet~\cite{ronneberger2015unet} architecture by cross-attention between controller features $\mathbf{x}_{C}$ and the modulated features $\mathbf{\hat{x}}$ stemming from DaLe.
\section{Future Work and Limitations}
The proposed DaAIR framework models degradation dependencies by allocating dedicated modeling capacity to individual tasks. Enhancing this framework with external inductive biases, such as edge information or frequency-related constraints, could improve the simultaneous handling of multiple degradation types, rather than relying solely on implicit degradation learning. Moreover, while this work utilizes synthetically degraded images, which have a significant domain gap from realistic scenarios, applying DaAIR to more complex, realistic degradation settings represents a promising direction for future research.
\section{Visual Results}
We provide additional visual results in \cref{fig:supp:visual_results} and feature visualizations in \cref{fig:supp:feat_visuals} to further underscore the strong restoration fidelity of our framework and the expressiveness of the learned features. These visualizations highlight the effectiveness of our approach in capturing intricate details and improving image quality across various degradation scenarios.

\end{document}